\documentclass[letterpaper, 10 pt, conference]{ieeeconf}

\IEEEoverridecommandlockouts

\usepackage{cite}
\usepackage{amsmath,amssymb,amsfonts}
\usepackage{algorithmic}
\usepackage{graphicx,subfigure}
\usepackage{textcomp}
\usepackage{xcolor}
\def\BibTeX{{\rm B\kern-.05em{\sc i\kern-.025em b}\kern-.08em
    T\kern-.1667em\lower.7ex\hbox{E}\kern-.125emX}}

\title{Efficient Minimal Solvers for Visual-Inertial Relative Pose Estimation in Multi-Camera Systems}
\author{Tao Li$^{1}$, Zhenbao Yu$^{2}$, Banglei Guan$^{2*}$, Jianli Han$^{1}$ and Weimin Lv$^{1}$
\thanks{
{Corresponding author: Banglei Guan.}} %
\thanks{$^{1}$Tao Li, Jianli Han, and Weimin Lv are with the College of Aerospace
Science and Engineering, Naval Aviation University, Yantai 264000, China
{\tt\footnotesize litao0931@alumni.nudt.edu.cn, jianlihan1585@163.com, 2016150315@jou.edu.cn}}
\thanks{$^{2} $Zhenbao Yu, and Banglei Guan is with the College of Aerospace Science and Engineering, National University of Defense Technology, Changsha 410000, China
{\tt\footnotesize zhenbaoyu@whu.edu.cn, guanbanglei12@nudt.edu.cn}}
}
\begin{document}
\maketitle
\begin{abstract}
Estimating the relative poses of multi-camera systems is a fundamental problem in computer vision, with critical applications in autonomous vehicles, mobile devices, and unmanned aerial vehicles (UAVs). However, existing solutions often suffer from high computational complexity or rely on an excessive number of point correspondences, limiting their real-world applicability. To address these limitations, we propose two efficient minimal solvers for estimating the relative poses of multi-camera systems using a novel parameterization. The first solver leverages the vertical direction prior provided by Inertial Measurement Units (IMUs), while the second utilizes the rotation axis direction prior from IMUs. Our methods require only four point correspondences and reduce the problem of multi-camera relative pose estimation to solving a univariate 6th-degree polynomial—a significant improvement over existing approaches, which typically involve 8th-degree polynomials. This reduction in computational complexity and correspondence requirements makes our solvers particularly effective when integrated into RANSAC frameworks, demonstrating strong potential for visual odometry applications. Through rigorous evaluations on synthetic data and the KITTI benchmark, our methods achieved superior computational efficiency and competitive accuracy compared to state-of-the-art algorithms.
\end{abstract}
\section{Introduction}
Relative pose estimation between two camera views is a core problem in computer vision, playing a key role in applications such as autonomous navigation, industrial inspection, and augmented reality \cite{campos2021orb, ding2025reposed, astermark2024fast, leiAMS2025, eichhardt2020relative}.
This problem has been extensively studied, with classical solutions including the normalized 8-point algorithm \cite{hartley2003multiple} and the 5-point algorithm \cite{nister2004efficient}. 
The accuracy, efficiency, and robustness of these algorithms are crucial, as they directly determine both the precision of visual localization and the quality of subsequent 3D reconstruction, ultimately affecting their real-world applicability.

\begin{figure}[t]             
  \vspace{-5pt}
     \centering    \includegraphics[width=0.9\linewidth]{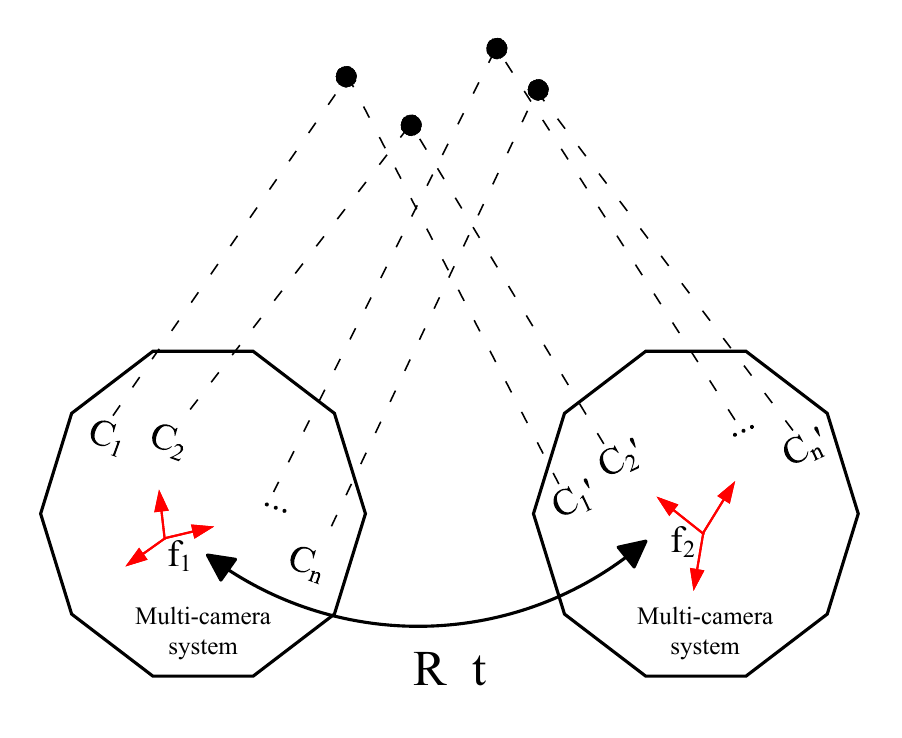}
    \centering
    \caption{The relative pose estimation for multi-camera systems.}
    \label{Multi-Camera}
    \vspace{-10pt}
\end{figure}

Based on the number of cameras used, relative pose estimation systems can be generally categorized into monocular \cite{Guan_TCYB2021, Zibin_TIP, TrifocalTensor2025} and multi-camera configurations \cite{guan2023minimal, guan2022relative_MM, guan2019minimal, tian2020efficient}.
Monocular systems utilize a single camera, while multi-camera systems employ multiple cameras rigidly mounted on a common platform.
Compared to monocular systems, multi-camera systems offer several distinct advantages. Primarily, they provide a wider field of view and richer environmental information, enabling synchronized observations from different viewpoints. Furthermore, by leveraging pre-calibrated baseline distances between cameras, multi-camera systems can directly recover absolute scene scale, thereby resolving the inherent scale ambiguity problem in monocular approaches. Moreover, multi-camera systems exhibit remarkable flexibility, as the number and orientation of cameras can be precisely tailored to suit specific scenes and tasks, enabling better adaptation to diverse challenging scenarios.

While multi-camera systems offer numerous advantages, they also introduce significant challenges. 
Unlike monocular systems, where light rays converge at a single center of projection, multi-camera configurations exhibit more complex geometric constraints due to their multiple projection centers \cite{pless2003using}, see Fig. \ref{Multi-Camera}. This inherent complexity makes multi-camera relative pose estimation considerably more intricate and challenging.
As a result, algorithms for multi-camera pose estimation typically require more feature correspondences and exhibit greater computational complexity. 
For instance, the linear method for multi-camera systems necessitates at least 17 point correspondences \cite{li2008linear}, a substantial increase compared to the 8 point correspondences required for monocular systems. 
When integrated with the RANSAC framework \cite{chum2005matching, barath2019progressive, barath2020magsac++}, this increased point correspondence requirement would further amplify the iteration count, thereby degrading computational efficiency and making it challenging to meet real-time requirements. 
Therefore, it is desirable to find more effective multi-camera solvers that can guarantee both high solution accuracy and computational efficiency.

 In this paper, we introduce two novel minimal solutions for this problem. The main contributions of this work are as follows:
\begin{itemize}
 \item  Using a novel depth-based parameterization for multi-camera system translation, we reformulate the new generalized epipolar constraint. 
 \item We introduce two novel minimal solvers using four point correspondences for multi-camera relative pose estimation. The first exploits the IMU vertical direction prior, while the second uses the IMU rotation axis direction prior. Both approaches reduce the problem to solving a 6th-degree univariate polynomial, offering significant computational advantages over existing 8th-degree solvers \cite{hee2014relative, liu2017robust, sweeney2014solving}.
 \item Extensive experiments demonstrate that our proposed solvers achieved competitive computational efficiency and accuracy compared to state-of-the-art methods.
\end{itemize}

\section{RELATED WORK}
For generalized camera pose estimation, Stewénius et al. introduced the 6-point minimal solver based on Gröbner-basis theory, which generates up to 64 potential solutions, resulting in significant computational demands \cite{henrikstewenius2005solutions}.
Guan et al. proposed a different set of 6-point minimal solvers, including both a general solver and specialized versions tailored for two-camera rig configurations. A key advancement lies in the incorporation of a ray bundle constraint, which significantly reduces the potential solution space \cite{guan2025six}.
In addition to the minimum solvers based on 6-point correspondences, Li et al. advanced the linear 17-point method by incorporating camera configuration effects on algorithm performance, resulting in enhanced capability to handle degenerate scenarios such as locally central and axial configurations \cite{li2008linear}.
Kim et al. systematically analyzed the degradation of the 17-point method through the decomposition of the measurement matrix into ray direction and projection center components \cite{kim2010degeneracy}. Chen et al. further demonstrated how visual overlap significantly impacts both the solvability and precision of the 17-point method, providing key insights for practical implementation \cite{xie202417}.
Complementary to minimal solvers depending on point correspondences, some approaches leveraging affine correspondences have also shown promising experimental performance \cite{guan2025affine, guan2021minimal}.  

Recent advances in generalized camera pose estimation have demonstrated the effectiveness of incorporating motion constraints \cite{guan2023minimal} or inertial sensors \cite{hee2014relative}. Lee et al. introduced a minimal 4-point algorithm for estimating the relative pose of multi-camera systems with known vertical direction \cite{hee2014relative}. Utilizing the hidden variable resultant method, their approach reduces the problem to solving an 8th-degree univariate polynomial, which can yield up to 8 real solutions. 
Beyond utilizing vertical direction information, Liu et al.'s 4-point algorithm using first-order rotation approximation achieves remarkable processing efficiency, making it highly suitable for real-time applications in dense urban driving conditions \cite{liu2017robust}.
Given the known rotation axis direction, Sweeney et al. developed an additional 4-point algorithm for relative pose estimation between generalized cameras \cite{sweeney2014solving}. More recently, Martyushev et al. introduced a minimal 5-point algorithm that leverages the known relative rotation angle \cite{martyushev2020efficient}.
These IMU-constrained approaches are particularly relevant to our work as they demonstrate how prior orientation information can significantly simplify the multi-camera problem while maintaining real-time performance.

\section{Generalized Camera Constraints under General Motion}
\subsection{Parameterization}
\begin{figure}[htbp]            
     \centering    \includegraphics[width=0.9\linewidth]{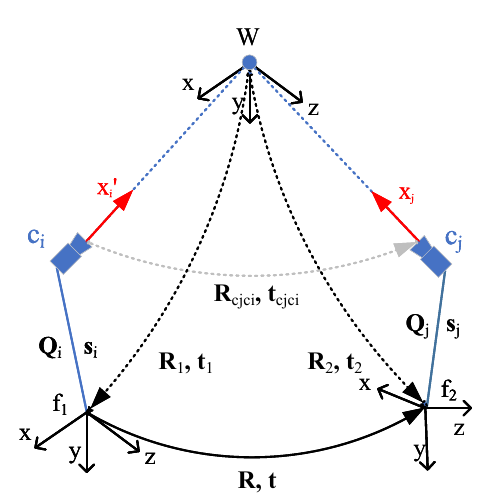}
    \centering
    \caption{The definition of coordinate systems in multi-camera systems.}
    \label{Coordinate}
\end{figure}

Consider a motion platform equipped with multiple cameras, also known as a multi-camera system, that moves from time $t_1$ to $t_2$. A 3D point $\mathbf{P}$ is captured by cameras $\mathbf{c}_i$ and $\mathbf{c}_j$ at different time instances. The motion platform's coordinate systems at these two different instances are denoted as $\mathbf{f}_1$ and $\mathbf{f}_2$, respectively. 
Following the recent work of \cite{guan2025affine}, the world coordinate system $\mathbf{W}$ is defined with its origin at point $\mathbf{P}$ and its orientation aligned with the platform's initial coordinate frame $\mathbf{f}_1$. Throughout this paper, $\mathbf{T_{ba}}$ represents the transformation matrix from frame $\mathbf{a}$ to frame $\mathbf{b}$. 
Specifically, the transformation matrix from frame $\mathbf{c}_i$ to frame $\mathbf{f}_1$ is $\mathbf{T}_{\mathbf{f}1\mathbf{c}i}=\left[ \begin{matrix} \mathbf{Q}_i & \mathbf{s}_i \\\end{matrix} \right]$, and the transformation matrix from frame $\mathbf{f}_1$ to frame $\mathbf{W}$ is $\mathbf{T}_{\mathbf{W}\mathbf{f}1}=\left[ \begin{matrix} \mathbf{R}_1 & \mathbf{t}_1 \\\end{matrix} \right]$. 
Similarly, the transformation matrix from frame $\mathbf{c}_j$ to frame $\mathbf{f}_2$ is $\mathbf{T}_{\mathbf{f}2\mathbf{c}j}=\left[ \begin{matrix} \mathbf{Q}_j & \mathbf{s}_j \\\end{matrix} \right]$, and the transformation matrix from frame $\mathbf{f}_2$ to frame $\mathbf{W}$ is $\mathbf{T}_{\mathbf{W}\mathbf{f}2}=\left[ \begin{matrix} \mathbf{R}_2 & \mathbf{t}_2 \\\end{matrix} \right]$. Since the orientation of the world coordinate system $\mathbf{W}$ aligns with the coordinate system $\mathbf{f}_1$, the rotation matrix $\mathbf{R}_1 = \mathbf{I}$. The transformation matrix from frame $\mathbf{f}_1$ to frame $\mathbf{f}_2$ is denoted as $\mathbf{T}_{\mathbf{f}2\mathbf{f}1}=\left[ \begin{matrix} \mathbf{R} & \mathbf{t}\\\end{matrix} \right]$, where $\mathbf{R}$ denotes the rotation matrix and $\mathbf{t}$ represents the translation vector. The specific meanings of each coordinate system, along with their corresponding matrix and vector representations,  are illustrated in Fig. \ref{Coordinate}.

Based on the coordinate system definitions above, the following transformation relationship holds:
\begin{equation}
	\begin{aligned}
   \mathbf{\bar{T}}_{\mathbf{f}2\mathbf{f}1}&={{\mathbf{\bar{T}}}_{\mathbf{f}2\mathbf{W}}}{{\mathbf{\bar{T}}}_{\mathbf{W}\mathbf{f}1}}=\left[ \begin{matrix}
   {{\mathbf{R}}_{2}} & {{\mathbf{t}}_{2}}  \\
   \mathbf{0} & 1  \\
\end{matrix} \right]\left[ \begin{matrix}
   \mathbf{I} & -{{\mathbf{t}}_{1}}  \\
   \mathbf{0} & 1  \\
\end{matrix} \right] \\ 
 & =\left[ \begin{matrix}
   {{\mathbf{R}}_{2}} & -{{\mathbf{R}}_{2}}{{\mathbf{t}}_{1}}+{{\mathbf{t}}_{2}}  \\
   \mathbf{0} & 1  \\
\end{matrix} \right]
        \end{aligned}.
	\label{a1}
\end{equation}
Here, $\mathbf{\bar{T}}$ denotes the homogeneous augmentation of the $3 \times 4$ matrix $\mathbf{T}$, obtained by appending the row vector $[0, 0, 0, 1]$. From Eq. \eqref{a1}, we obtain:
\begin{equation}
	\begin{aligned}
   & \mathbf{R}={{\mathbf{R}}_{2}} \\ 
  & \mathbf{t}=-{{\mathbf{R}}_{2}}{{\mathbf{t}}_{1}}+{{\mathbf{t}}_{2}} \\ 
        \end{aligned}.
	\label{a2}
\end{equation}

In the generalized camera model, the point correspondence ${{\mathbf{u}}_{i}}\leftrightarrow {{\mathbf{u}}_{j}}$ of the 3D point $\mathbf{P}$ can be viewed as the intersection of two Plücker lines ${{\mathbf{L}}_{i}}={{\left[   \begin{matrix}   {{\mathbf{x}}_{i}}^{\mathrm{T}} & {{\mathbf{q}}_{i}}^{\mathrm{T}}  \\\end{matrix} \right]}^{\mathrm{T}}}$ and ${{\mathbf{L}}_{j}}={{\left[   \begin{matrix}   {{\mathbf{x}}_{j}}^{\mathrm{T}} & {{\mathbf{q}}_{j}}^{\mathrm{T}}  \\\end{matrix} \right]}^{\mathrm{T}}}$. In the Plücker line representation, ${{\mathbf{x}}_{*}}$ denotes the unit direction vector, given by $\mathbf{x}_{*} = (\mathbf{Q}_{*} *\mathbf{u}_{*})/\begin{Vmatrix}\mathbf{Q}_{*} *\mathbf{u}_{*}\end{Vmatrix}$, and ${{\mathbf{q}}_{*}}={{\mathbf{s}}_{*}}\times {{\mathbf{x}}_{*}}$ denotes the moment vector. According to Plücker line geometry, any point $\mathbf{U}(\lambda )$ on the line can be parameterized as:
\begin{equation}
	\begin{aligned}
\mathbf{U}(\lambda )={{\mathbf{x}}}\times {{\mathbf{q}}}+{{\lambda }}{{\mathbf{x}}},\lambda \in \mathbb{R}
\end{aligned}.
	\label{a3}
\end{equation}

Since the 3D point $\mathbf{P}$ lies at the intersection of two Plücker lines, its coordinates can be parameterized by these two lines simultaneously. The corresponding coordinates of point $\mathbf{P}$ along these two lines are respectively given by $\mathbf{U}(\lambda_1 )$ and $\mathbf{U}(\lambda_2 )$. Specifically, $\mathbf{U}(\lambda_1 )$ and $\mathbf{U}(\lambda_2 )$ represent the coordinates of $\mathbf{P}$ within the motion platform's coordinate systems $\mathbf{f}_1$ and $\mathbf{f}_2$, respectively, where $\mathbf{P}$ corresponds to the origin $[0, 0, 0]^{\mathrm{T}}$ of the world coordinate system $\mathbf{W}$. Therefore, the following relationships hold:
\begin{equation}
	\begin{aligned}
\mathbf{U}(\lambda_1 )&={{\mathbf{x}}_{1}}\times {{\mathbf{q}}_{1}}+{{\lambda }_{1}}{{\mathbf{x}}_{1}}={{\mathbf{R}}_{1}}{[0, 0, 0]^{\mathrm{T}}}+{{\mathbf{t}}_{1}}\\
\mathbf{U}(\lambda_2 )&={{\mathbf{x}}_{2}}\times {{\mathbf{q}}_{2}}+{{\lambda }_{2}}{{\mathbf{x}}_{2}}={{\mathbf{R}}_{2}}{[0, 0, 0]^{\mathrm{T}}}+{{\mathbf{t}}_{2}}
\end{aligned}.
	\label{a4}
\end{equation}
Then, the expressions of $\mathbf{t}_{1}$ and $\mathbf{t}_{2}$ can be derived using the depths $\lambda_1$ and $\lambda_2$.
\begin{equation}
	\begin{aligned}
{{\mathbf{t}}_{1}}&={{\mathbf{x}}_{1}}\times {{\mathbf{q}}_{1}}+{{\lambda }_{1}}{{\mathbf{x}}_{1}}\\
{{\mathbf{t}}_{2}}&={{\mathbf{x}}_{2}}\times {{\mathbf{q}}_{2}}+{{\lambda }_{2}}{{\mathbf{x}}_{2}}
\end{aligned}.
	\label{a5}
\end{equation}
This shows that the translation vector $\mathbf{t}$ can be expressed using two depth parameters $\lambda_1$ and $\lambda_2$, rather than adopting the conventional ${[t_x, t_y, t_z]}^{\mathrm{T}}$ representation. Utilizing this novel parameterization, we can derive new formulations for the generalized epipolar constraint (GEC).

\subsection{\label{sec:GEC}Generalized Epipolar Constraint}
Since ${{\mathbf{L}}_{i}}$ and ${{\mathbf{L}}_{j}}$ represent Plücker lines in two distinct coordinate frames $\mathbf{f}_1$ and $\mathbf{f}_2$ respectively, they need to be transformed into a common coordinate system. When transforming the Plücker line ${{\mathbf{L}}_{i}}$ from frame $\mathbf{f}_1$ to frame $\mathbf{f}_2$, the resulting coordinate are given by $\mathbf{L}_{i}^{\prime}={{\left[ \begin{matrix}{{\mathbf{x}}_{i}}{{^{\prime }}^{\mathrm{T}}} & {{\mathbf{q}}_{i}}{{^{\prime }}^{\mathrm{T}}}  \\\end{matrix} \right]}^{\mathrm{T}}}$. The relationship between $\mathbf{L}_{i}^{\prime}$ and ${{\mathbf{L}}_{i}}$ is described by the following transformation:
\begin{equation}
	\begin{aligned} \mathbf{L}_{i}^{\prime}=\left[ \begin{matrix}
   \mathbf{R} & \mathbf{0}  \\
   {{[\mathbf{t}]}_{\times }}\mathbf{R} & \mathbf{R}  \\
\end{matrix} \right]{{\mathbf{L}}_{i}}=\left[ \begin{matrix}
   \mathbf{R}{{\mathbf{x}}_{i}}  \\
   \mathbf{R}{{\mathbf{q}}_{i}}+{{[\mathbf{t}]}_{\times }}\mathbf{R}{{\mathbf{x}}_{i}}  \\
\end{matrix} \right].
\end{aligned}
	\label{b0}
\end{equation}

According to\cite{pless2003using}, the necessary and sufficient condition for two straight lines $\mathbf{L}_{i}^{\prime}$ and ${{\mathbf{L}}_{j}}$ to intersect within the same coordinate system $\mathbf{f}_2$ is given by:
\begin{equation}
	\begin{aligned}{{\mathbf{x}}_{i}}{{^{\prime }}^{\mathrm{T}}}{{\mathbf{q}}_{j}}+{{\mathbf{x}}_{j}}^{\mathrm{T}}{{\mathbf{q}}_{i}}^{\prime }=0
\end{aligned}.
	\label{b1}
\end{equation}
By substituting \eqref{b0} into \eqref{b1}, the generalized epipolar constraint can be derived as follows:
\begin{equation}
	\begin{aligned}{(\mathbf{R}{{\mathbf{x}}_{i}})}^{\mathrm{T}}{{\mathbf{q}}_{j}}+{{{\mathbf{x}}_{j}}^{\mathrm{T}}}\mathbf{R}{{\mathbf{q}}_{i}}+{{{\mathbf{x}}_{j}}^{\mathrm{T}}}{{[\mathbf{t}]}_{\times }}\mathbf{R}{{\mathbf{x}}_{i}}=0.
    \end{aligned}
	\label{b2}
\end{equation}
Substituting \eqref{a2} into \eqref{b2} yields:
\begin{equation}
	\begin{aligned}
{{\mathbf{x}}_{i}}^{\mathrm{T}}{{\mathbf{R}}^{\mathrm{T}}}{{\mathbf{q}}_{j}}+{{\mathbf{x}}_{j}}^{\mathrm{T}}\mathbf{R}{{\mathbf{q}}_{i}}+{{\mathbf{x}}_{j}}^{\mathrm{T}}({{[{{\mathbf{t}}_{2}}]}_{\times }}\mathbf{R}-\mathbf{R}{{[{{\mathbf{t}}_{1}}]}_{\times }}){{\mathbf{x}}_{i}}=0
    \end{aligned}.
	\label{b3}
\end{equation}
Finally, by substituting \eqref{a5} into \eqref{b3}, the generalized epipolar constraints formulated in terms of depths $\lambda_1$ and $\lambda_2$ can be derived:
\begin{equation}
	\begin{aligned}
      & -{{\lambda }_{1}}{{\mathbf{x}}_{j}}^{\mathrm{T}}\mathbf{R}{{[{{\mathbf{x}}_{1}}]}_{\times }}{{\mathbf{x}}_{i}}+{{\lambda }_{2}}{{\mathbf{x}}_{j}}^{\mathrm{T}}({{[{{\mathbf{x}}_{2}}]}_{\times }}\mathbf{R}){{\mathbf{x}}_{i}}+{{\mathbf{x}}_{j}}^{\mathrm{T}}\mathbf{R}{{\mathbf{q}}_{i}} +  \\ 
 & {{\mathbf{x}}_{i}}^{\mathrm{T}}{{\mathbf{R}}^{\mathrm{T}}}{{\mathbf{q}}_{j}}+{{\mathbf{x}}_{j}}^{\mathrm{T}}({{[{{\mathbf{x}}_{2}}\times {{\mathbf{q}}_{2}}]}_{\times }}\mathbf{R}-\mathbf{R}{{[{{\mathbf{x}}_{1}}\times {{\mathbf{q}}_{1}}]}_{\times }}){{\mathbf{x}}_{i}}=0 \\ 
    \end{aligned}.
	\label{b4}
\end{equation}

\section{4-Point algorithm for Multi-Camera Systems}
\subsection{\label{sec:4pt_vertical}4-Point algorithm with Known Vertical Direction}  
Through IMU alignment, the platform's y-axis can be precisely aligned with gravitational direction, ensuring orthogonality to the ground plane.
Let $\mathbf{R}_{imu1}$ and $\mathbf{R}_{imu2}$ denote the alignment matrices derived from IMU measurements at timestamps $t_1$ and $t_2$, respectively. The relationship between the original rotation matrix $\mathbf{R}$ and its aliened counterpart ${{\mathbf{R}}_{\mathbf{v}2\mathbf{v}1}}$ is given by:
\begin{equation}
	\begin{aligned}
{{\mathbf{R}}}={{\mathbf{R}}_{imu2}}^{\mathrm{T}}{{\mathbf{R}}_{\mathbf{v}2\mathbf{v}1}}{{\mathbf{R}}_{imu1}}
\end{aligned}.
	\label{d0}
\end{equation}
Then, the problem can be reduced to solving for $\mathbf{R}_{\mathbf{v}2\mathbf{v}1}$, which depends solely on the yaw angle. Using Cayley parameterization,  the aligned rotation matrix $\mathbf{R}_{\mathbf{v}2\mathbf{v}1}$ takes the form:  
\begin{equation}
	\begin{aligned}
\mathbf{R}_{\mathbf{v}2\mathbf{v}1}=\frac{1}{1+{{s}^{2}}}\left[ \begin{matrix}
   1-{{s}^{2}} & 0 & 2s  \\
   0 & 1+{{s}^{2}} & 0  \\
   -2s & 0 & 1-{{s}^{2}}  \\
\end{matrix} \right]
\end{aligned},
	\label{d1}
\end{equation}
where $\theta_y$ denotes the yaw angle and $s=\tan(\theta_y/2)$.

Substituting \eqref{d0} into the generalized epipolar constraint \eqref{b4}, we can obtain:
\begin{equation}
	\begin{aligned}
      & -{{\lambda }_{1}}{{\mathbf{\tilde{x}}}_{j}}^{\mathrm{T}}\mathbf{R}_{\mathbf{v}2\mathbf{v}1}{{[{{\mathbf{\tilde{x}}}_{1}}]}_{\times }}{{\mathbf{\tilde{x}}}_{i}}+{{\lambda }_{2}}{{\mathbf{\tilde{x}}}_{j}}^{\mathrm{T}}({{[{{\mathbf{\tilde{x}}}_{2}}]}_{\times }}\mathbf{R}_{\mathbf{v}2\mathbf{v}1}){{\mathbf{\tilde{x}}}_{i}}\\ 
 &+{{\mathbf{\tilde{x}}}_{j}}^{\mathrm{T}}\mathbf{R}_{\mathbf{v}2\mathbf{v}1}{{\mathbf{\tilde{q}}}_{i}} +{{\mathbf{\tilde{x}}}_{i}}^{\mathrm{T}}{{\mathbf{R}_{\mathbf{v}2\mathbf{v}1}}^{\mathrm{T}}}{{\mathbf{\tilde{q}}}_{j}}\\ 
 &+{{\mathbf{\tilde{x}}}_{j}}^{\mathrm{T}}({{[{{\mathbf{\tilde{x}}}_{2}}\times {{\mathbf{\tilde{q}}}_{2}}]}_{\times }}\mathbf{R}_{\mathbf{v}2\mathbf{v}1}-\mathbf{R}_{\mathbf{v}2\mathbf{v}1}{{[{{\mathbf{\tilde{x}}}_{1}}\times {{\mathbf{\tilde{q}}}_{1}}]}_{\times }}){{\mathbf{\tilde{x}}}_{i}}=0 \\ 
    \end{aligned}.
	\label{d2}
\end{equation}
Comparing \eqref{d2} with \eqref{b4} reveals their structural similarity, with the key distinction being that the variables in \eqref{d2} have been adjusted by alignment matrices. The detailed adjustments include:
${{\mathbf{\tilde{x}}}_{i}}={{\mathbf{R}}_{imu1}}{{\mathbf{x}}_{i}}$, ${{\mathbf{\tilde{x}}}_{j}}={{\mathbf{R}}_{imu2}}{{\mathbf{x}}_{j}}$, ${{\mathbf{\tilde{q}}}_{i}}={{\mathbf{R}}_{imu1}}{{\mathbf{q}}_{i}}$, ${{\mathbf{\tilde{q}}}_{j}}={{\mathbf{R}}_{imu2}}{{\mathbf{q}}_{j}}$.

Using three non-origin point correspondences, we derive three equations from \eqref{d2}. We specifically exclude the point correspondence corresponding to the world origin $\mathbf{P}$ because its constraint equation reduces to 0, providing no valuable information for solving the unknown parameters. The three remaining equations can be organized in matrix form as:
\begin{equation}
	\begin{aligned}
{{\mathbf{F}}_{3\times 3}}(s)\left[ \begin{matrix}
   {{\lambda }_{1}}  \\
   {{\lambda }_{2}}  \\
   1  \\
\end{matrix} \right]={{\mathbf{0}}_{3\times 1}}
\end{aligned}.
	\label{d4}
\end{equation}
Since the system of equations \eqref{d4} has a non-trivial solution, the matrix $\mathbf{F}(s)$ must satisfy $\text{rank}(\mathbf{F}(s)) \leq 2$, which implies:
\begin{equation}
	\begin{aligned}
\text{det}({\mathbf{F}}(s))= 0
\end{aligned}.
	\label{d5}
\end{equation}
Based on \eqref{d5}, we can obtain a 6th-degree univariate polynomial in the variable $s$, providing up to 6 real solutions.

Once the solutions for $s$ are obtained, they can be substituted back into \eqref{d1} and subsequently into \eqref{d0} to recover the complete rotation matrix $\mathbf{R}$.

\subsection{\label{sec:4pt_axis}4-Point algorithm with Known Rotation Axis Direction}
Instead of using the Cayley parameterization or Euler angles to represent rotations, this section employs the quaternion representation. The rotation quaternion is denoted as ${{\left[ \begin{matrix}
   \begin{matrix}
   {{q}_{1}} & {{q}_{2}} & {{q}_{3}} & \omega  \\
\end{matrix}  \\
\end{matrix} \right]}^{\mathrm{T}}}$, which can also be compactly written as ${{\left[ \begin{matrix}
   \begin{matrix}
   {{\mathbf{v}}^{\mathrm{T}}} & \omega  \\
\end{matrix}  \\
\end{matrix} \right]}^{\mathrm{T}}}$, where $\mathbf{v} = [q_1, q_2, q_3]^{\mathrm{T}}$ is the vector part and $\omega$ is the scalar part. For a unit quaternion, the vector component $\mathbf{v}$ defines the rotation axis direction, and the scalar component $\omega$ gives the cosine of half the rotation angle.
The relationship between a unit quaternion and its corresponding rotation matrix is given by:
\begin{equation}
	\begin{aligned}
{{\mathbf{R}}}= \mathbf{v}{{\mathbf{v}}^{\top }}+{{\omega}^{2}}\mathbf{I}+2\omega{{[\mathbf{v}]}_{\times }}+{{[\mathbf{v}]}_{\times }}^{2}
\end{aligned},
	\label{g1}
\end{equation}
where $[\mathbf{v}]_{\times}$ denotes the skew-symmetric matrix of vector $\mathbf{v}$.

When the quaternion representing $\mathbf{R}$ is not necessarily normalized, its vector and scalar components are denoted as $\mathbf{\tilde{v}}$ and $\tilde{\omega}$ respectively. Here, $\tilde{\mathbf{v}} = \mathbf{v}/||\mathbf{v}||$ is the unit vector along the rotation axis, and $\tilde{\omega} = \omega/||\mathbf{v}||$. 
The rotation matrix $\tilde{\mathbf{R}}$ constructed using these non-normalized components follows the same formula structure and is proportional to the true rotation matrix $\mathbf{R}$:
\begin{equation}
	\begin{aligned}
{{\mathbf{R}}}\sim {\mathbf{\tilde{R}}}= \mathbf{\tilde{v}}{{\mathbf{\tilde{v}}}^{\top }}+{{\tilde{\omega}}^{2}}\mathbf{I}+2\tilde{\omega}{{[\mathbf{\tilde{v}}]}_{\times }}+{{[\mathbf{\tilde{v}}]}_{\times }}^{2}
\end{aligned}.
	\label{g2}
\end{equation}
Note that $\tilde{\mathbf{R}}$ and $\mathbf{R}$ differ only by a scale factor, which does not affect the solution validity since the generalized epipolar constraint is a homogeneous equation.

 Substituting the quaternion-based rotation representation from \eqref{g2} into the generalized epipolar constraint \eqref{b4}, we can establish a system of three equations:
 \begin{equation}
	\begin{aligned}
{{\mathbf{F}}_{3\times 3}}(\tilde{\omega})\left[ \begin{matrix}
   {{\lambda }_{1}}  \\
   {{\lambda }_{2}}  \\
   1  \\
\end{matrix} \right]={{\mathbf{0}}_{3\times 1}}
\end{aligned}.
	\label{g3}
\end{equation}
 Analogous to the derivation of \eqref{d5}, this system leads to a similar determinant condition:
\begin{equation}
	\begin{aligned}
\text{det}({\mathbf{F}}(\tilde{\omega}))= 0
\end{aligned}.
	\label{g4}
\end{equation}
Consistent with the previous finding for the variable $s$, expanding \eqref{g4} yields a 6th-degree univariate polynomial in the variable $\tilde{\omega}$, which provides up to 6 real solutions.

The obtained solutions for $\tilde{\omega}$, combined with the assumed known rotation axis $\tilde{\mathbf{v}}$, define an unnormalized quaternion. By normalizing this quaternion, we recover the unit vector $\mathbf{v}$ and scalar $\omega$ components of the unit quaternion. Subsequently, substituting these normalized components into equation \eqref{g1} enables the complete recovery of the rotation matrix $\mathbf{R}$.

\subsection{Translation Vector Estimation}
Once the values for $s$ or $\tilde{\omega}$ are determined from the 6th-degree polynomial, they are substituted into the corresponding matrix $\mathbf{F}(\cdot)$ defined in \eqref{d4} or \eqref{g3}. We can obtain a homogeneous linear equation system in terms of the unknown depth parameters $\lambda_1$ and $\lambda_2$. Using Singular Value Decomposition (SVD), the depth parameters $\lambda_1$ and $\lambda_2$ are obtained from the null space of the matrix $\mathbf{F}(\cdot)$.

Subsequently, by substituting these calculated depths $\lambda_1$ and $\lambda_2$ back into the definitions of $\mathbf{t}_1$ and $\mathbf{t}_2$ as provided in \eqref{a5}, we can compute the specific values for $\mathbf{t}_1$ and $\mathbf{t}_2$. Finally, the overall translation vector $\mathbf{t}$ can be recovered using the following formula:
\begin{equation}
	\begin{aligned} 
  \mathbf{t}=-{{\mathbf{R}}}{{\mathbf{t}}_{1}}+{{\mathbf{t}}_{2}} \\ 
        \end{aligned},
	\label{g5}
\end{equation}
where $\mathbf{R}$ corresponds to the rotation matrix determined in the previous steps through either the Cayley parameterization or quaternion representation.

\subsection{\label{sec:Degeneracy}Degeneracy Analysis}
Prior work \cite{hee2014relative} has identified that degeneracy occurs when a multi-camera system undergoes pure translation while relying solely on intra-camera correspondences. Although our approach employs the depth-based parameterization for translation representation rather than the conventional $[t_x, t_y, t_z]^{\mathrm{T}}$ formulation, the multi-camera system still exhibits degeneracy under pure translational motion with intra-camera correspondences, resulting in unrecoverable scale factors \cite{guan2025affine}.
Consider the case of pure translation where the relative rotation $\mathbf{R} = \mathbf{I}$, combined with intra-camera correspondences where $\mathbf{s}_i = \mathbf{s}_j$. Under these conditions, the cross-term in the generalized epipolar constraint vanishes:
\begin{equation}
	\begin{aligned}
{{\mathbf{x}}_{i}}^{\text{T}}{{\mathbf{R}}}^{\text{T}}{{\mathbf{q}}_{j}}+{{\mathbf{x}}_{j}}^{\text{T}}{{\mathbf{R}}}{{\mathbf{q}}_{i}}={{\mathbf{x}}_{i}}^{\text{T}}{{\mathbf{s}}_{i}}\times {{\mathbf{x}}_{j}}+{{\mathbf{x}}_{j}}^{\text{T}}{{\mathbf{s}}_{i}}\times {{\mathbf{x}}_{i}}=\mathbf{0}
    \end{aligned}.
	\label{k1}
\end{equation}
Consequently, the generalized epipolar constraint \eqref{b3} reduces to the following form:
\begin{equation}
	\begin{aligned}
{{\mathbf{x}}_{j}}^{\mathrm{T}}({{[{{\mathbf{t}}_{2}-\mathbf{t}}_{1}}]}_{\times }){{\mathbf{x}}_{i}}=0
    \end{aligned}.
	\label{k2}
\end{equation}
In this degenerate case, the constraint matrix becomes homogeneous with respect to the relative translation $\mathbf{t}_2 - \mathbf{t}_1$. Consequently, introducing a free parameter $\kappa$ reveals that $\kappa(\mathbf{t}_2 - \mathbf{t}_1)$ always satisfies equation \eqref{k2}, indicating the loss of scale information.

\section{Experiment}
In the experiments, the solver presented in Section~\ref{sec:4pt_vertical} is denoted as \texttt{4pt-Our}, while the solver from Section~\ref{sec:4pt_axis} is referred to as \texttt{4pt-Our-Axis}.
For comprehensive comparison, we benchmark our solvers against both state-of-the-art 4-point methods, including \texttt{4pt-Lee}~\cite{hee2014relative}, \texttt{4pt-Liu}~\cite{liu2017robust}, and \texttt{4pt-Sweeney}~\cite{sweeney2014solving}, as well as the recently proposed 5-point method \texttt{5pt-Martyushev}~\cite{martyushev2020efficient}. 
In the experiments, we evaluate rotation and translation errors using the following metrics~\cite{guan2025affine}:
 \begin{equation}
	\begin{aligned} 
\varepsilon_{\mathbf{R}}&=\arccos((\mathrm{trace}(\mathbf{R}_{gt}\mathbf{R}^T)-1)/2) \\  
\varepsilon_{\mathbf{t}}&=\arccos((\mathbf{t}_{gt}^T\mathbf{t})/(\|\mathbf{t}_{gt}\|\cdot\|\mathbf{t}\|))
\end{aligned}.
\label{erro_metrics}
\end{equation}
Here, $\mathbf{R}_{gt}$ and $\mathbf{t}_{gt}$ represent the ground truth values, while $\mathbf{R}$ and $\mathbf{t}$ represent the estimated values. $\varepsilon_{\mathbf{R}}$ represents the angular error of the rotation, $\varepsilon_{\mathbf{t}}$ indicates the angular error of the translation.

\subsection{Efficiency Comparison and Numerical Stability}

\begin{table*}[htbp]
	\caption{Runtime comparison of relative pose estimation solvers (unit:~$\mu s$).}
	\begin{center}
		\setlength{\tabcolsep}{2.0mm}{
			\scalebox{1.1}{
				\begin{tabular}{|c||c|c|c|c|c|c|c|}
					\hline					
					Methods       & 5pt-Martyushev \cite{martyushev2020efficient} & 4pt-Lee \cite{hee2014relative}   &4pt-Liu \cite{liu2017robust}       & 4pt-Sweeney \cite{sweeney2014solving}& 4pt-Our    & 4pt-Our-Axis \\
					\hline
					Runtime  & 591.356 & 28.024 & \textbf{8.268} & 22.186 & 21.343 & 14.334 \\
					\hline
		\end{tabular}}}
	\end{center}
	\label{tab:SolverTime_4pt}
\end{table*}

\begin{figure}[htbp]
	\begin{center}
		\subfigure[$\varepsilon_{\mathbf{R}}$]
		{
			\includegraphics[width=0.45\linewidth]{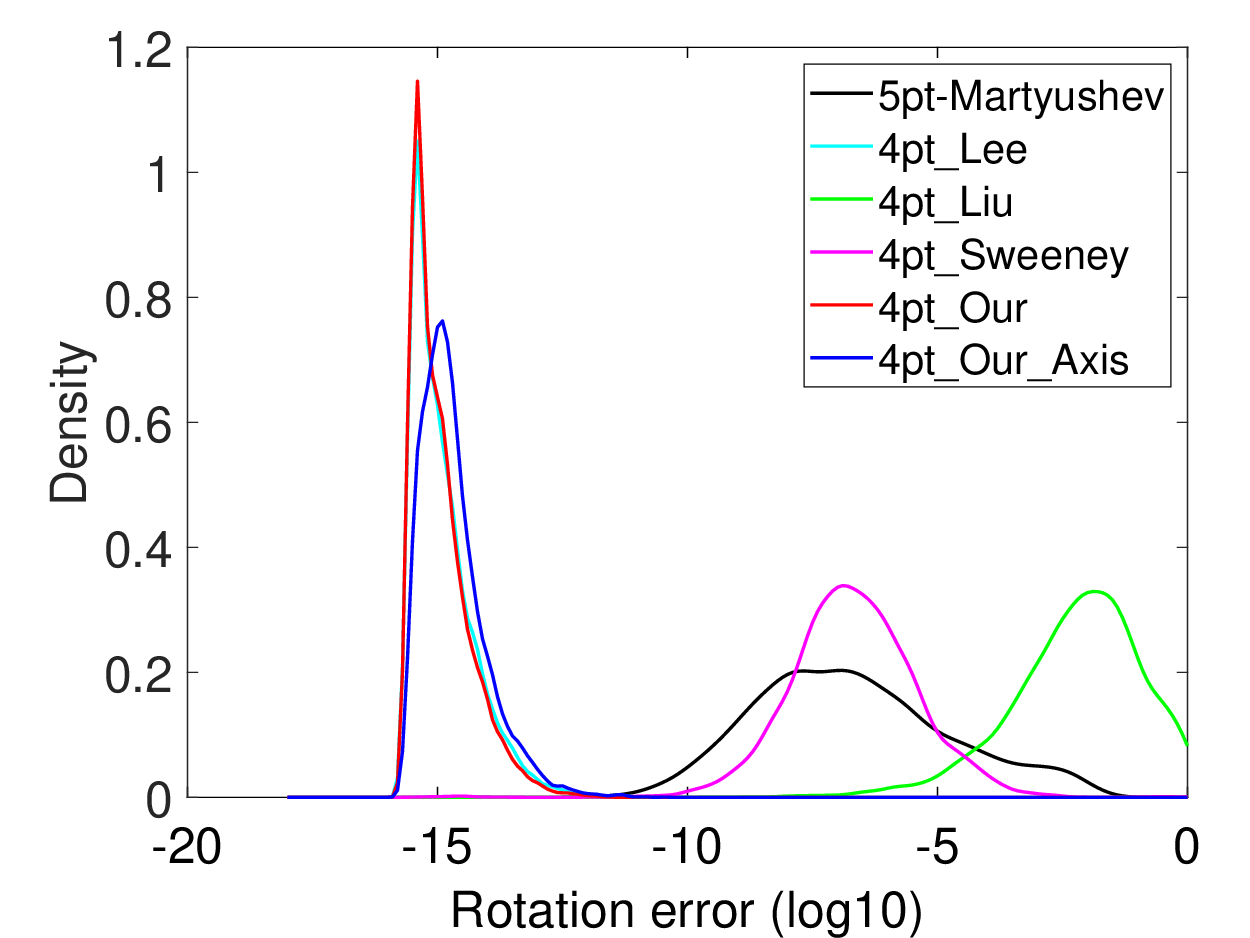}
		}
		\subfigure[$\varepsilon_{\mathbf{t}}$]
		{
			\includegraphics[width=0.45\linewidth]{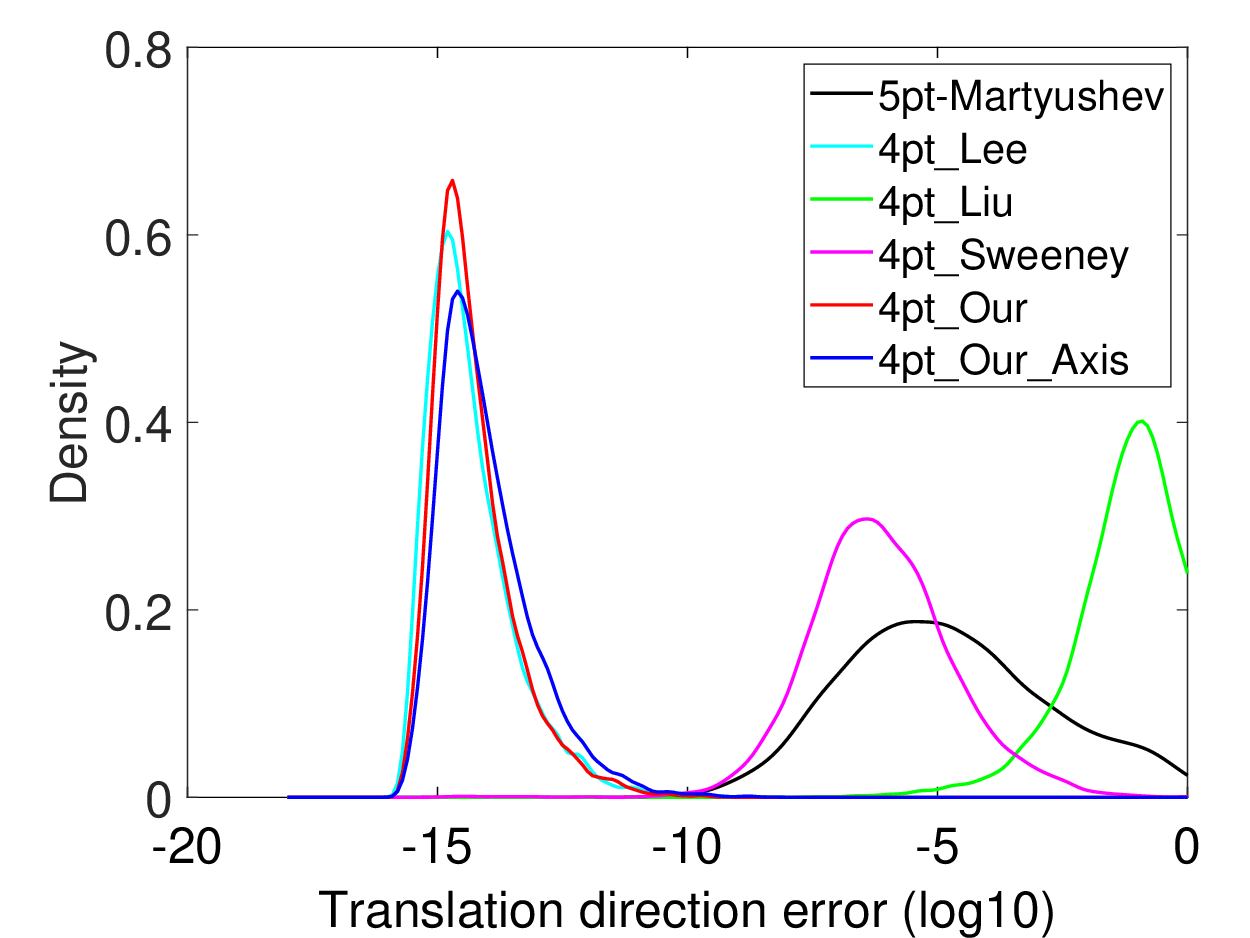}
		}
	\end{center}
	\caption{Numerical stability comparison of relative pose estimation solvers.}
	\label{fig:Numerical}
\end{figure}

To ensure a fair comparison, all solvers were evaluated on an AMD R9-7945HX 2.50 GHz processor using C++ implementations. 
Table~\ref{tab:SolverTime_4pt} presents the average computation times across 10,000 independent runs. 
The experimental results indicate that all 4-point solvers are substantially more computationally efficient than the 5-point solver \texttt{5pt-Martyushev}~\cite{martyushev2020efficient}.
Among them, \texttt{4pt-Liu}~\cite{liu2017robust} achieves the fastest runtime due to its use of a first-order rotation approximation to simplify the computation. Among solvers that do not rely on such approximations, our proposed \texttt{4pt-Our} and \texttt{4pt-Our-Axis} methods run more efficiently than \texttt{4pt-Lee}~\cite{hee2014relative} and \texttt{4pt-Sweeney}~\cite{sweeney2014solving}. This efficiency gain stems from our derivation of more compact polynomial equations with fewer terms, which reduces the computational burden of polynomial expansion and root-solving. 
Overall, these results demonstrate the strong computational efficiency of our proposed solvers.

Fig.~\ref{fig:Numerical} presents a comparative analysis of solver numerical stability under noise-free conditions. Each solver was executed 10,000 independent runs, with the probability density functions plotted against the $\log_{10}$ of the rotation errors and translation errors. A higher and more left-shifted peak in these distributions indicates superior numerical stability, whereas a broader and right-shifted distribution suggests poorer stability.
Among solvers utilizing IMU vertical angle priors, our proposed \texttt{4pt-Our} achieves exceptional numerical stability, with errors concentrated around $10^{-15}$.  
The \texttt{4pt-Lee}~\cite{hee2014relative} solver shows nearly comparable performance, whereas \texttt{4pt-Liu}~\cite{liu2017robust} exhibits degraded stability due to its reliance on first-order rotation approximation.
Similarly, for solvers using IMU rotation axis priors, our \texttt{4pt-Our-Axis} method shows better performance, with errors concentrated around $10^{-14}$.
 
\subsection{Experiments on Synthetic Data}
For synthetic evaluation, we adopted a multi-camera system consisting of two simulated cameras with a fixed baseline of 1 meter. 
The system's position and orientation varied randomly over time, with translational displacements limited to a maximum of 3 meters and rotational variations confined to a range of $-10^{\circ}$ to $10^{\circ}$. 
The camera intrinsics were configured with a focal length of 400 pixels and principal point coordinates centered at (320, 240).
Spatially distributed 3D points were randomly generated and designed to be simultaneously observable by both cameras. Point correspondences were exclusively intra-camera correspondences.
The evaluation comprised 1,000 independent trials, with pose estimation accuracy quantified using median rotation and translation errors.

\subsubsection{\label{sec:imagenoise}Accuracy with Image Noise} 
\begin{figure}[htbp]
	\begin{center}
		\includegraphics[width=0.9\linewidth]{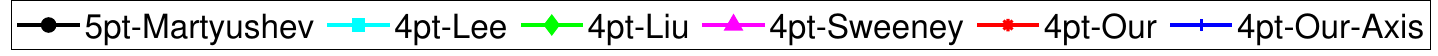}\\
		\subfigure[${\varepsilon_{\bf{R}}}$]
		{
			\includegraphics[width=0.45\linewidth]{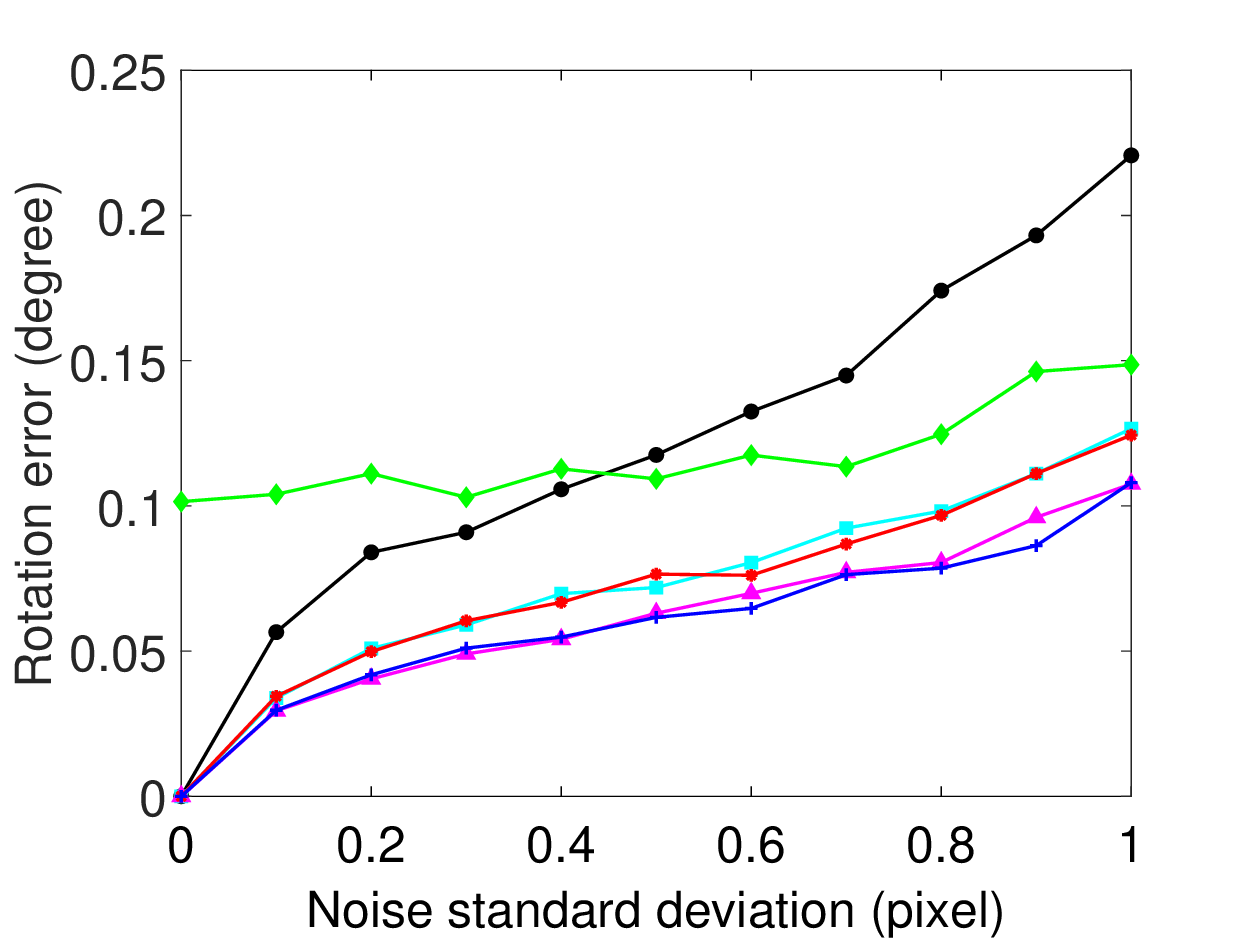}
		}
 \subfigure[$\varepsilon_{\mathbf{t}}$]
		{
			\includegraphics[width=0.45\linewidth]{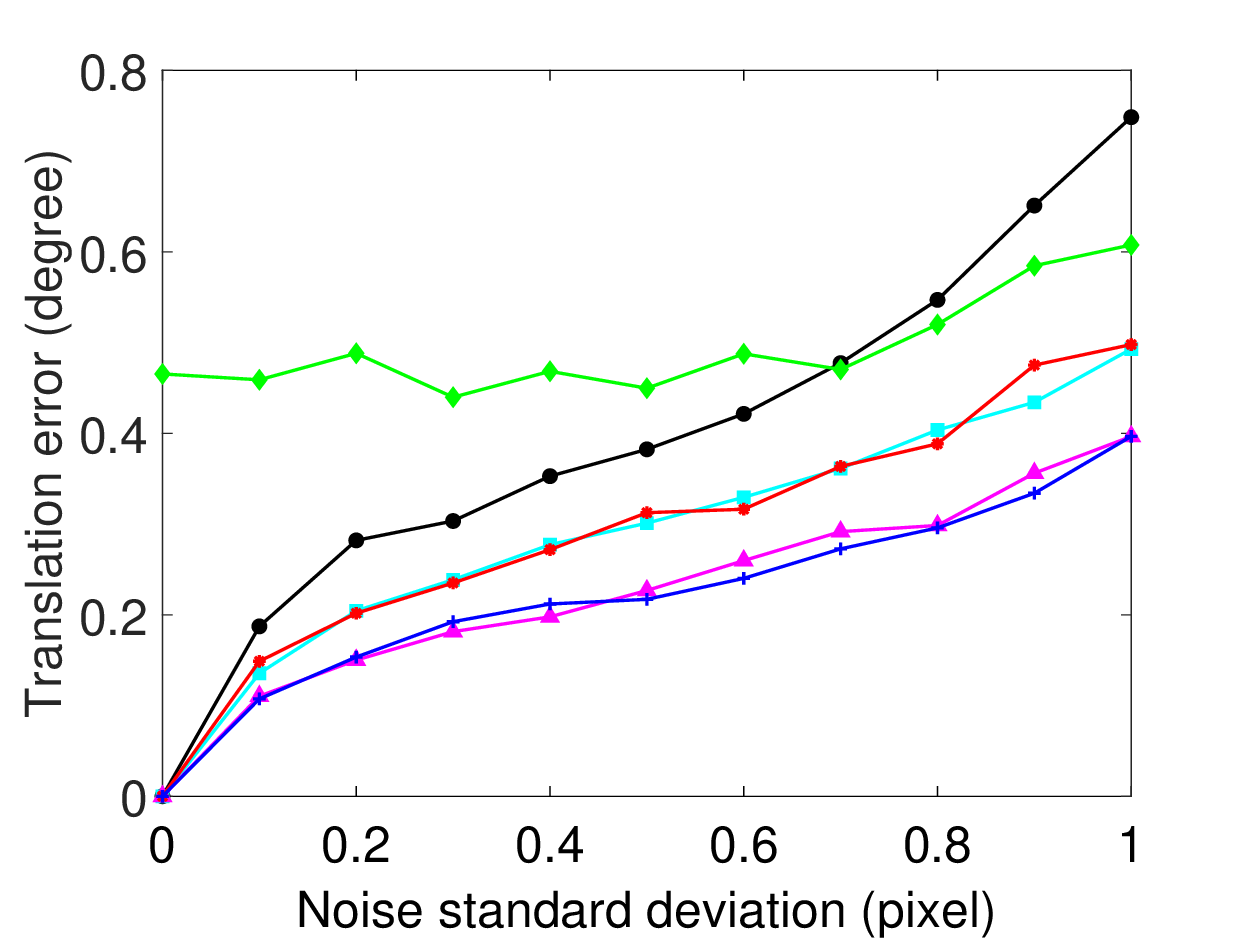}
		}

		\subfigure[${\varepsilon_{\bf{R}}}$]
		{
			\includegraphics[width=0.45\linewidth]{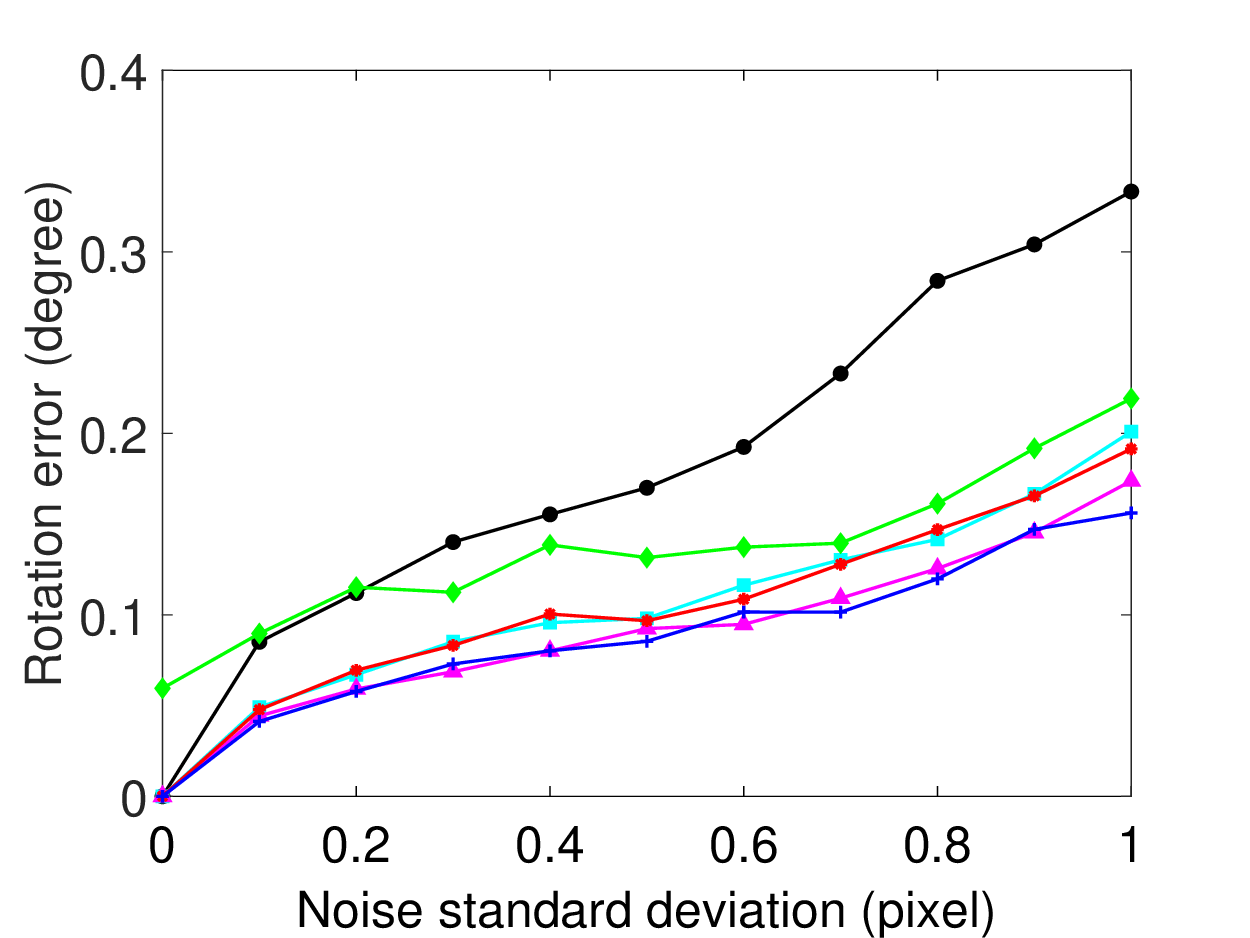}
		}
\subfigure[$\varepsilon_{\mathbf{t}}$]
		{
			\includegraphics[width=0.45\linewidth]{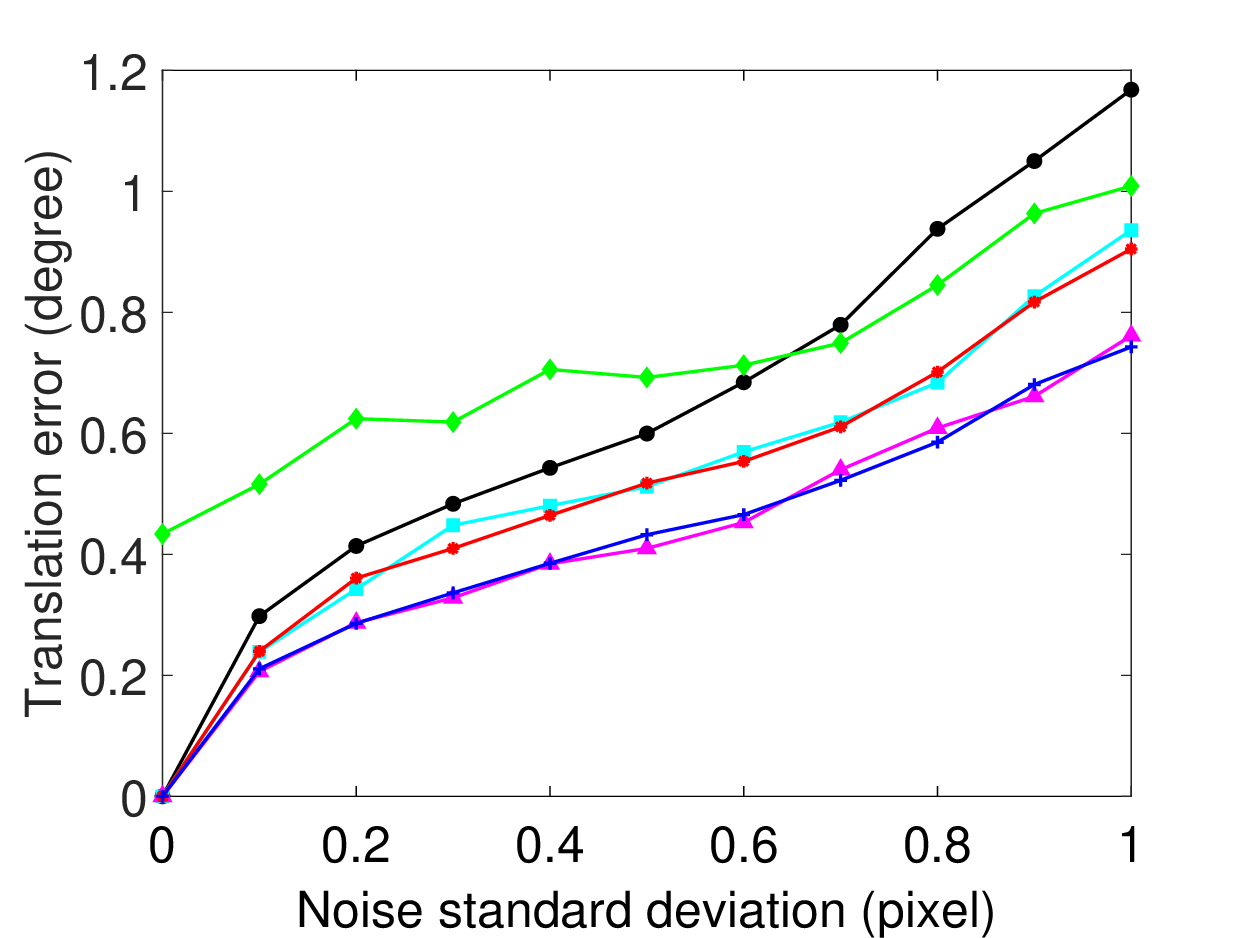}
		}
 
		\subfigure[${\varepsilon_{\bf{R}}}$]
		{
			\includegraphics[width=0.45\linewidth]{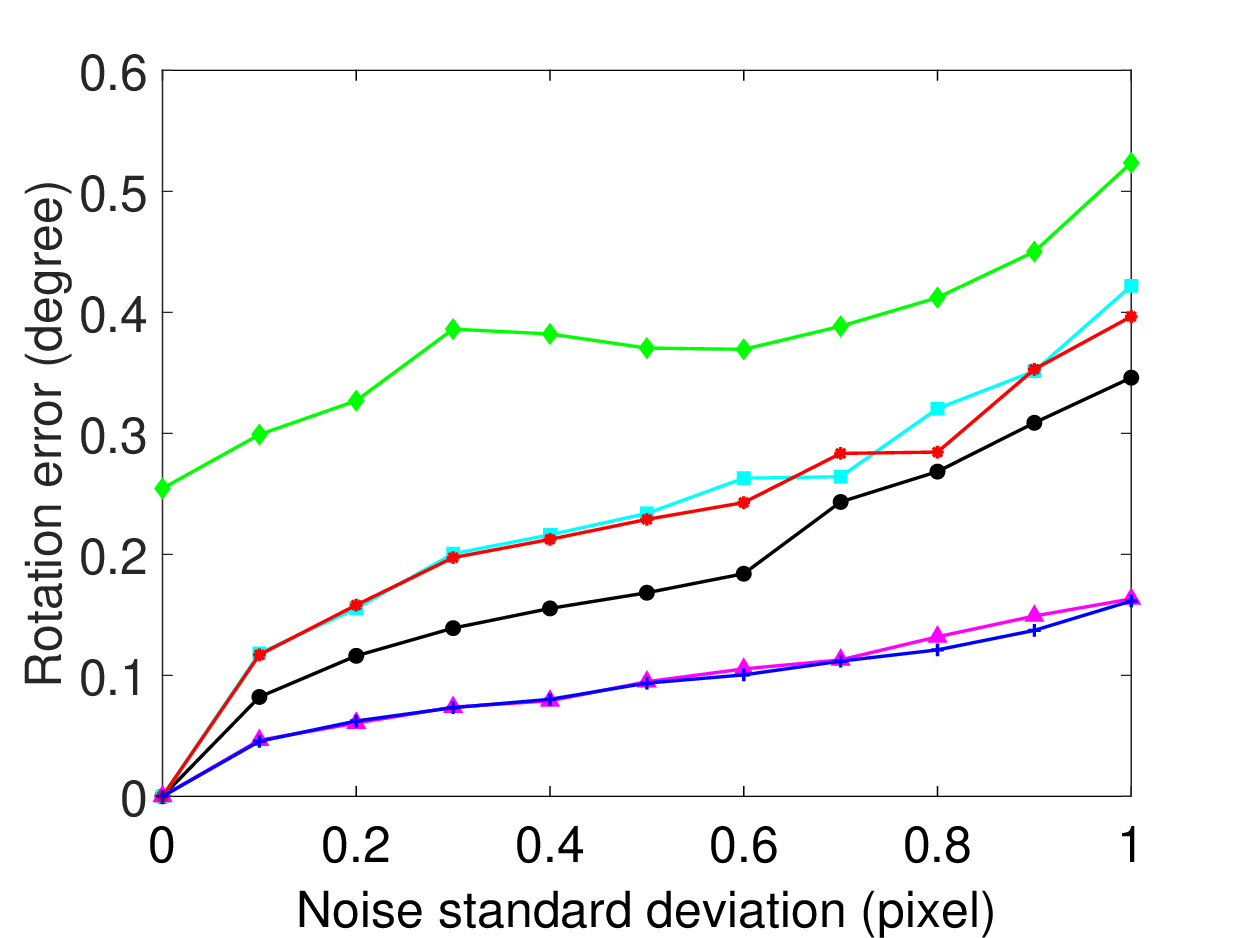}
		}
\subfigure[$\varepsilon_{\mathbf{t}}$]
		{
			\includegraphics[width=0.45\linewidth]{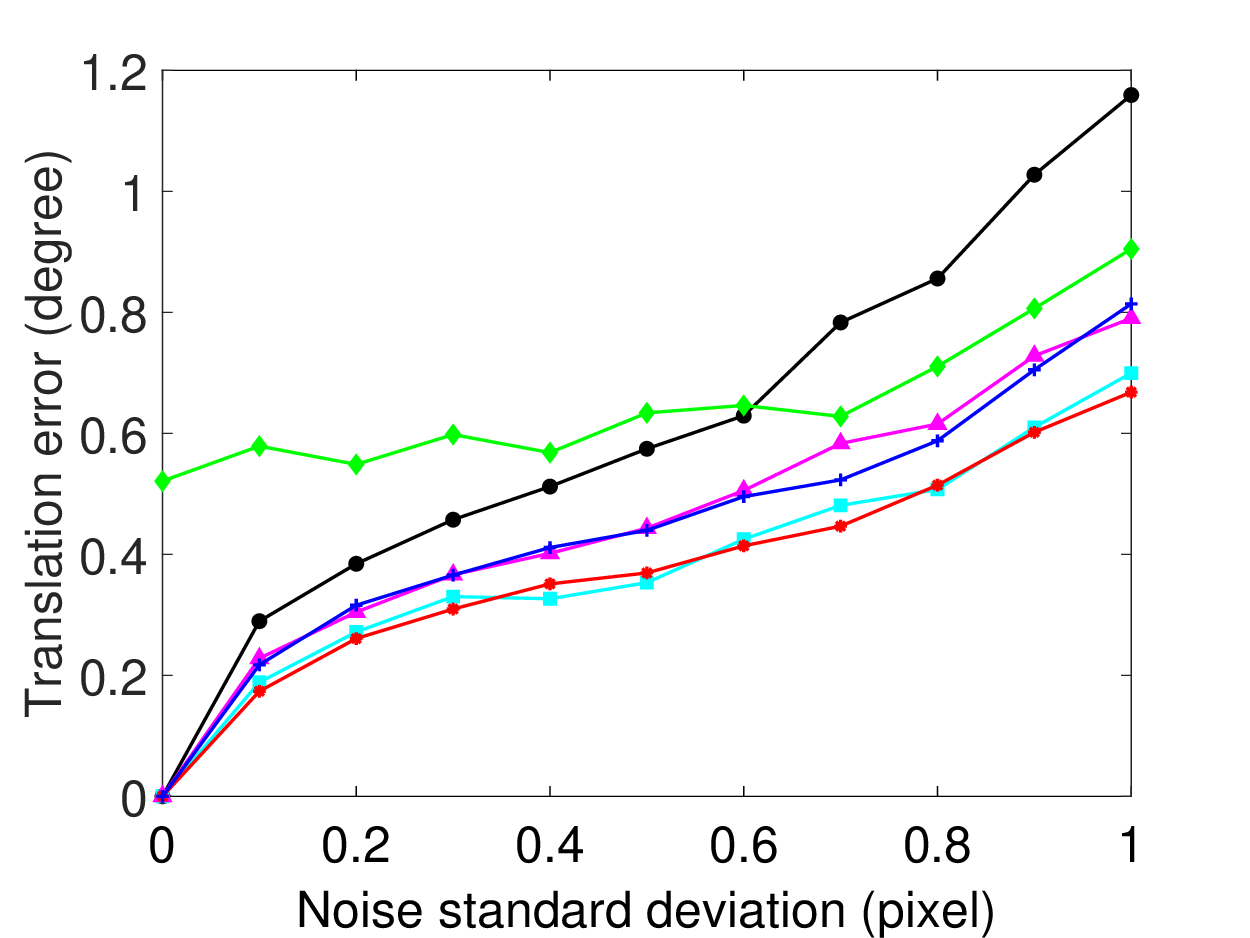}
		}
	\end{center}
	\caption{Rotation and translation errors for multi-camera systems with increasing image noise. The top, middle, and bottom rows present the performance of different solvers under forward, random, and sideways motions, respectively.}
	\label{fig:image_noise}
\end{figure}

We evaluate the accuracy of the solvers in the presence of image noise by introducing Gaussian noise with standard deviations ranging from 0 to 1.0 pixels. The multi-camera system motion patterns included forward, random, and sideways motions. As shown in Fig.~\ref{fig:image_noise}, the 4-point solvers generally demonstrate greater robustness to image noise compared to the 5-point solver \texttt{5pt-Martyushev}~\cite{martyushev2020efficient}, except in the case of sideways motion.
For the 4-point solvers leveraging IMU rotation axis priors, both our proposed \texttt{4pt-Our-Axis} and \texttt{4pt-Sweeney}~\cite{sweeney2014solving} achieve comparable accuracy and outperform all other evaluated methods in most experimental scenarios. For the 4-point solvers utilizing IMU vertical angle priors, our \texttt{4pt-Our} and \texttt{4pt-Lee}~\cite{hee2014relative} exhibited similar accuracy levels,  while both showed significant accuracy improvements over the \texttt{4pt-Liu}~\cite{liu2017robust} method. This performance disparity primarily stems from \texttt{4pt-Liu}'s reliance on first-order rotation approximation, which is only effective for small-angle rotations and becomes insufficient when handling larger angular motions.
\subsubsection{\label{sec:IMUnoise} Accuracy with IMU Noise}
In practical applications, high-end IMUs can achieve angle measurement accuracy as precise as $0.02^\circ$,  while even low-cost IMUs typically offer errors below $0.5^\circ$~\cite{kukelova2010closed}. 
To thoroughly evaluate our algorithm's robustness against IMU measurement errors, we conducted simulations with angular noise spanning $0^\circ$ to $1^\circ$—a range that encompasses the performance of both commercial and industrial-grade IMUs. 
The evaluated methods differ in their use of IMU priors. \texttt{4pt-Lee}~\cite{hee2014relative}, \texttt{4pt-Liu}~\cite{liu2017robust}, and \texttt{4pt-Our} rely on IMU vertical angle priors, while \texttt{4pt-Our-Axis} and \texttt{4pt-Sweeney}~\cite{sweeney2014solving} leverage IMU rotation axis priors. 
 To ensure consistency in our simulations, we generated noisy data based on IMU vertical angle noise, as this can be directly transformed into equivalent IMU rotation axis noise. Specifically, noise was introduced by perturbing the pitch and roll angles, while the image noise was kept constant at 0.5 pixels. 

Since the \texttt{5pt-Martyushev}~\cite{martyushev2020efficient} method depends on the rotation angle rather than the vertical angle, the comparison of the impact of pitch and roll angle errors is restricted to the \texttt{4pt-Lee}~\cite{hee2014relative}, \texttt{4pt-Liu}~\cite{liu2017robust}, and \texttt{4pt-Sweeney}~\cite{sweeney2014solving} methods. 

\begin{figure}[htbp]
	\begin{center}
				\includegraphics[width=0.9\linewidth]{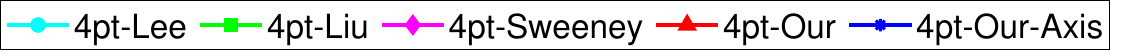}\\
		\subfigure[${\varepsilon_{\bf{R}}}$]
		{
			\includegraphics[width=0.45\linewidth]{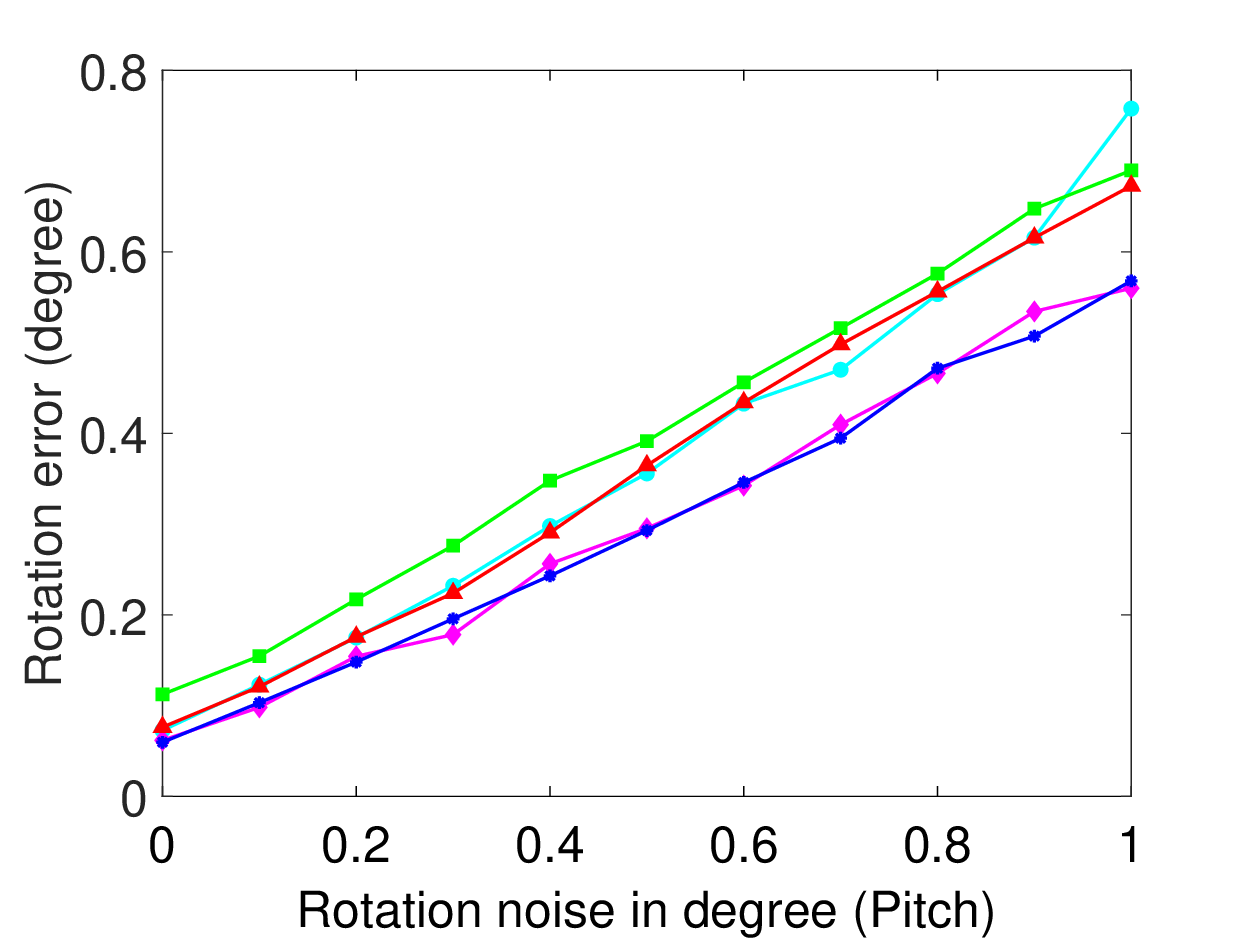}
		}
	\subfigure[$\varepsilon_{\mathbf{t}}$]
		{
			\includegraphics[width=0.45\linewidth]{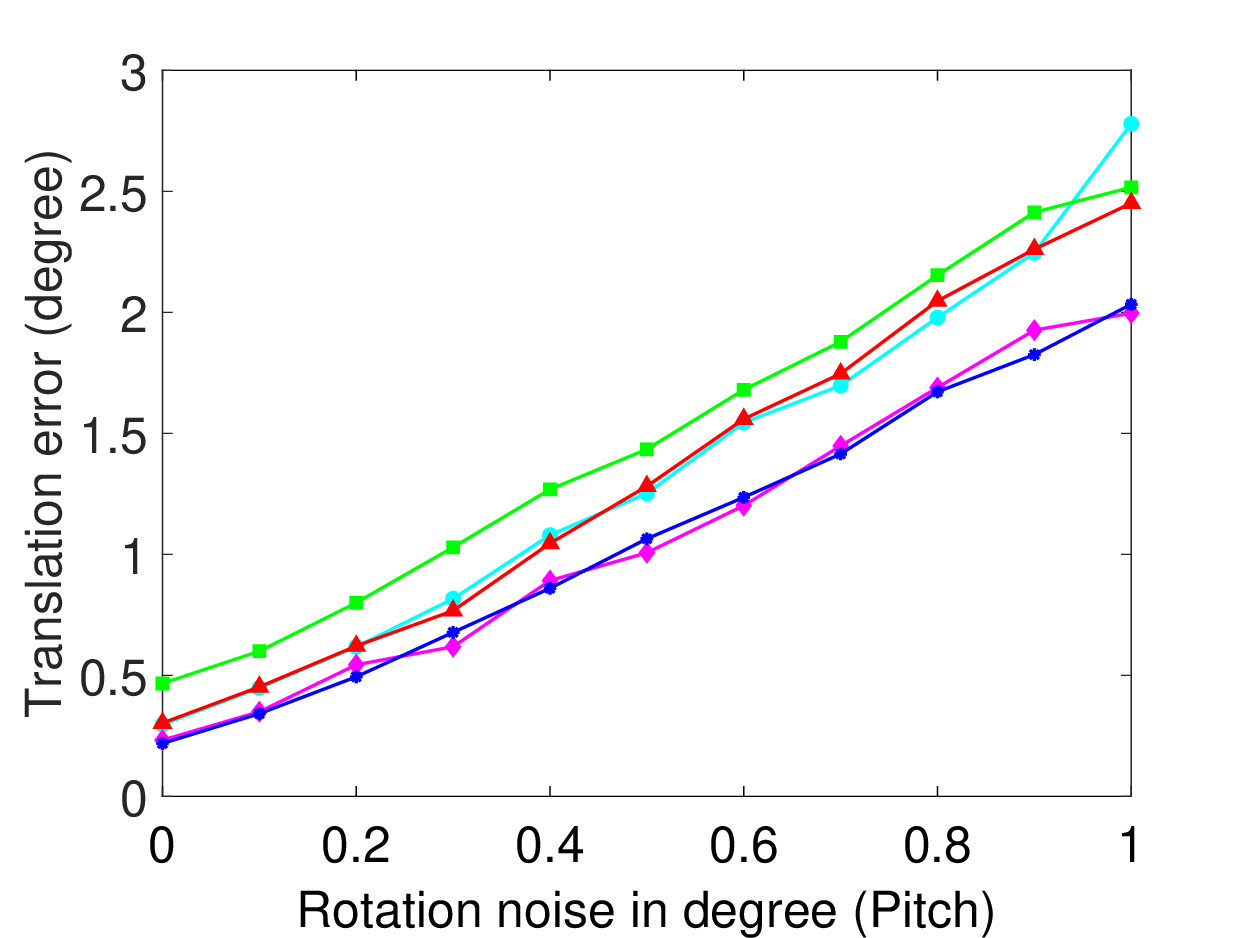}
		}
 
		\subfigure[${\varepsilon_{\bf{R}}}$]
		{
			\includegraphics[width=0.45\linewidth]{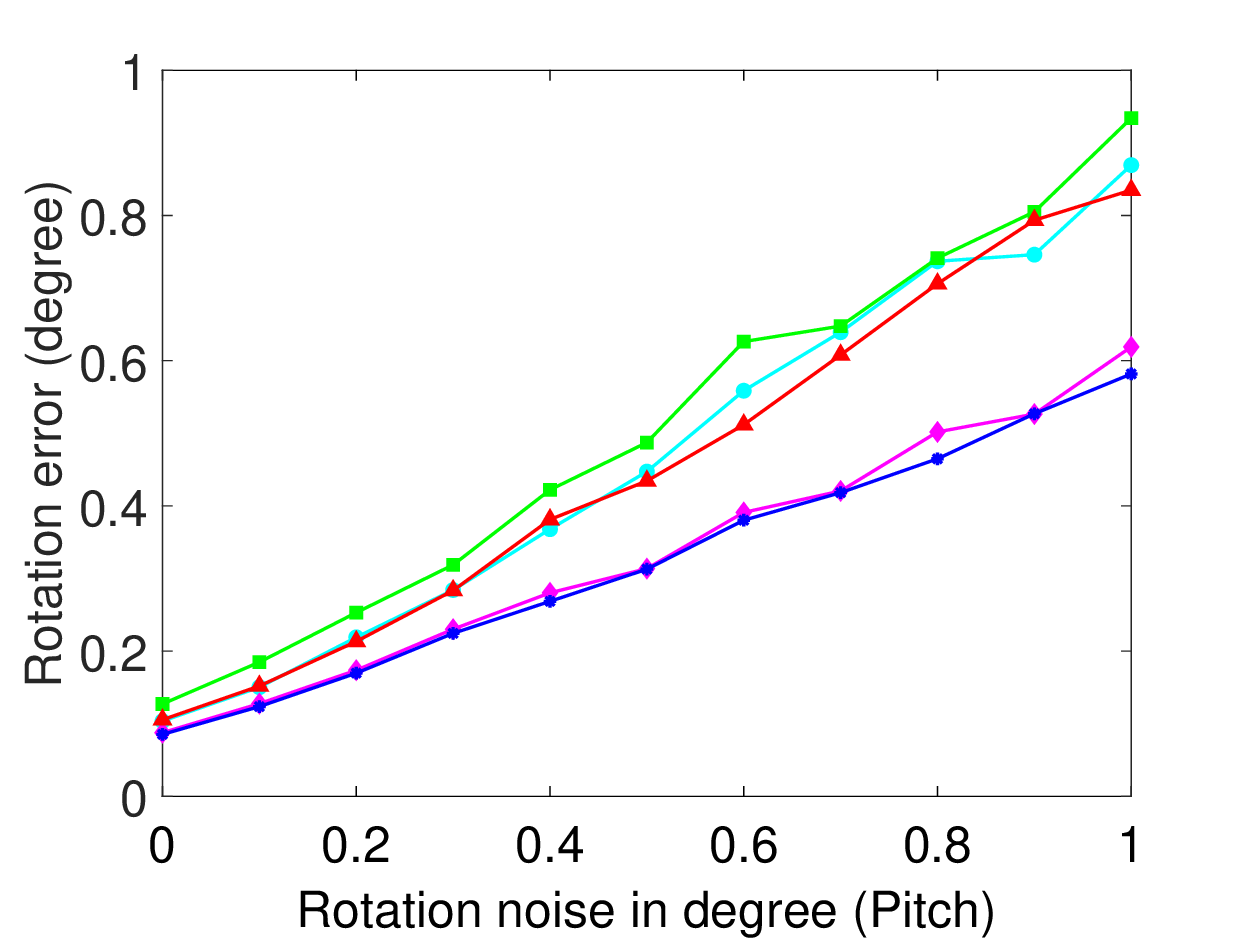}
		}
	\subfigure[$\varepsilon_{\mathbf{t}}$]
		{
			\includegraphics[width=0.45\linewidth]{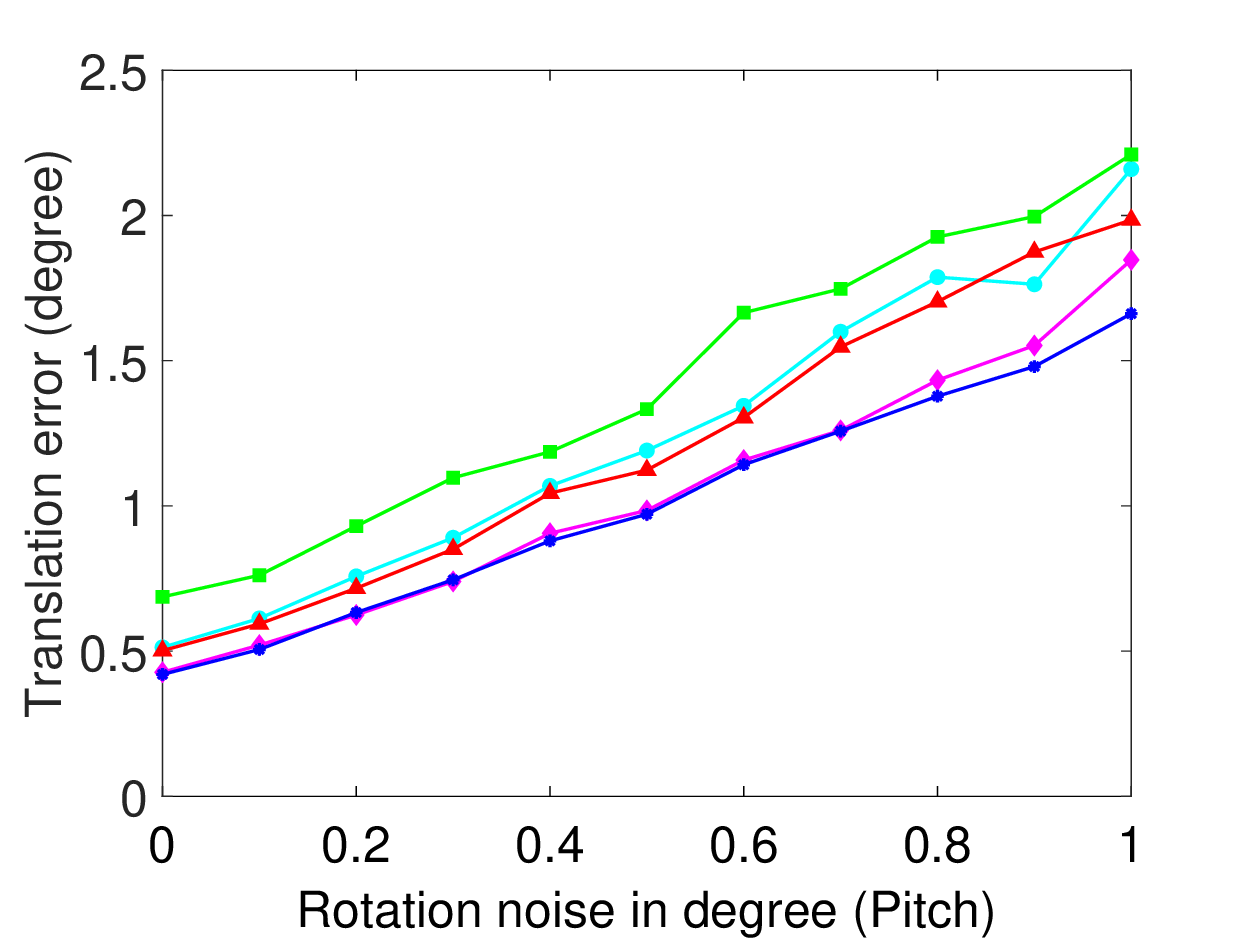}
		}
	 
		\subfigure[${\varepsilon_{\bf{R}}}$]
		{
			\includegraphics[width=0.45\linewidth]{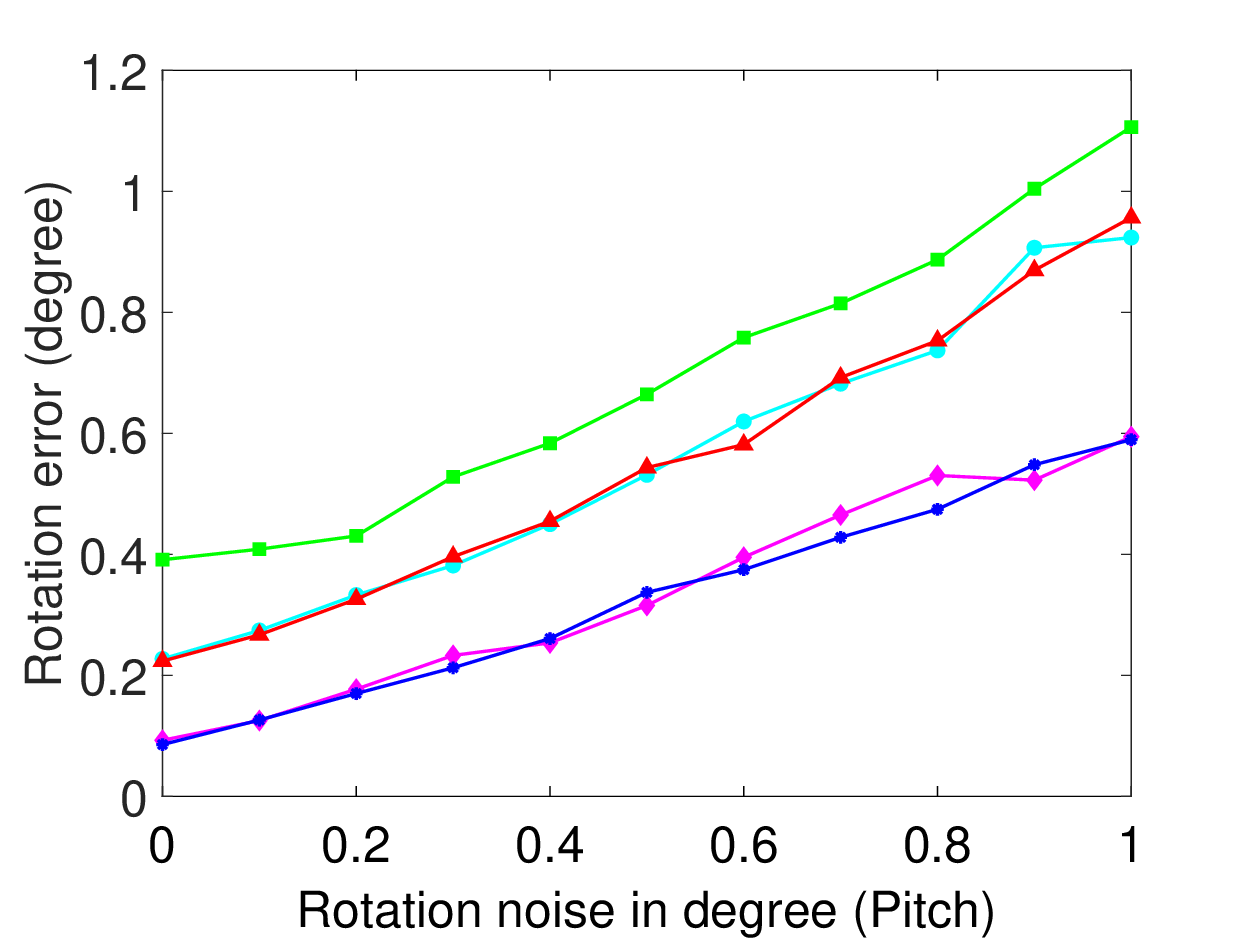}
		}
	\subfigure[$\varepsilon_{\mathbf{t}}$]
		{
			\includegraphics[width=0.45\linewidth]{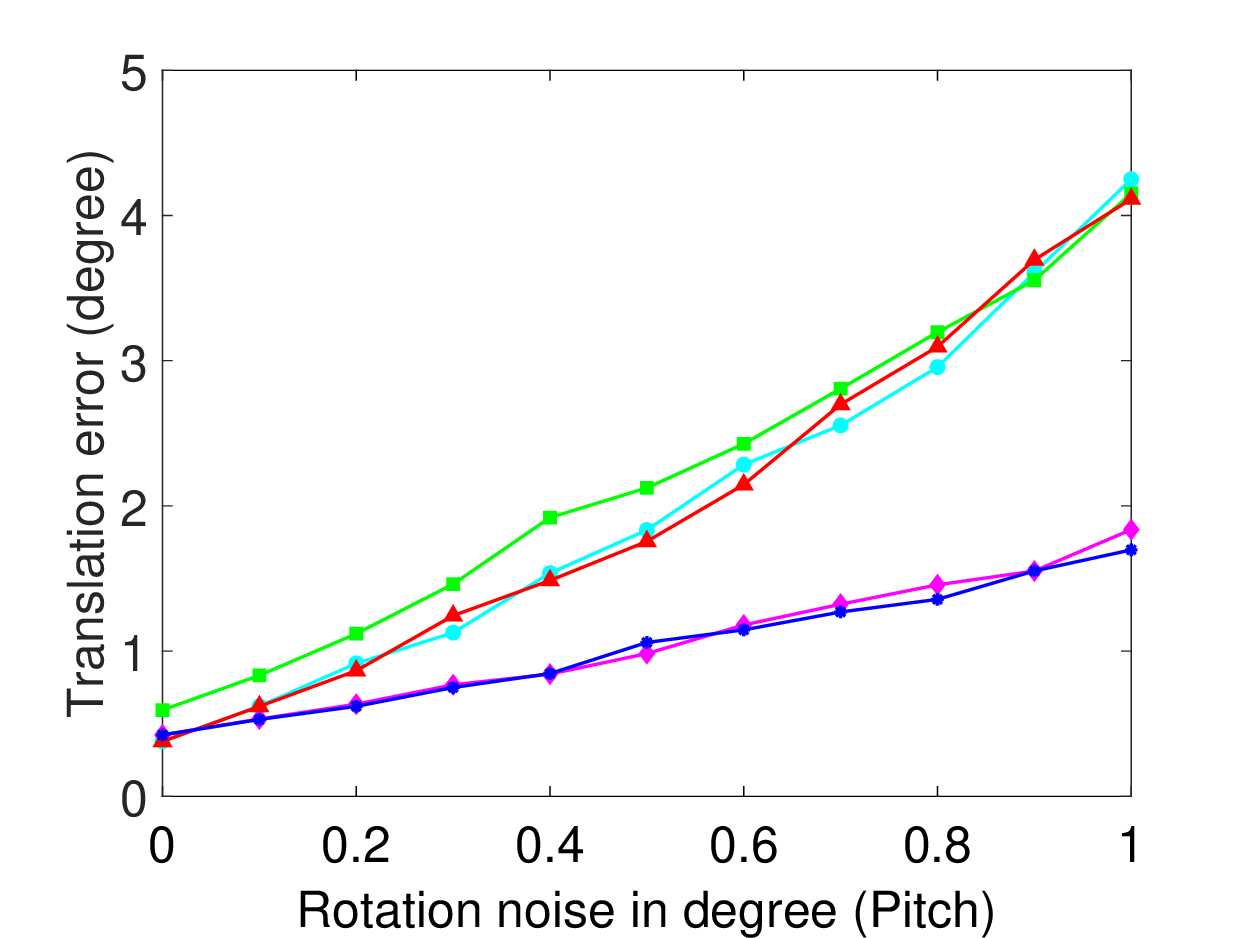}
		}
	\end{center}
	\caption{Rotation and translation errors for multi-camera systems with increasing IMU pitch angle noise. The top, middle, and bottom rows present the performance of different solvers under forward, random, and sideways motions, respectively.}
	\label{fig:pitch_angle_noise}
\end{figure}

\begin{figure}[htbp]
	\begin{center}
				\includegraphics[width=0.9\linewidth]{fig/simulation/legend3.pdf}\\
		\subfigure[${\varepsilon_{\bf{R}}}$]
		{
			\includegraphics[width=0.45\linewidth]{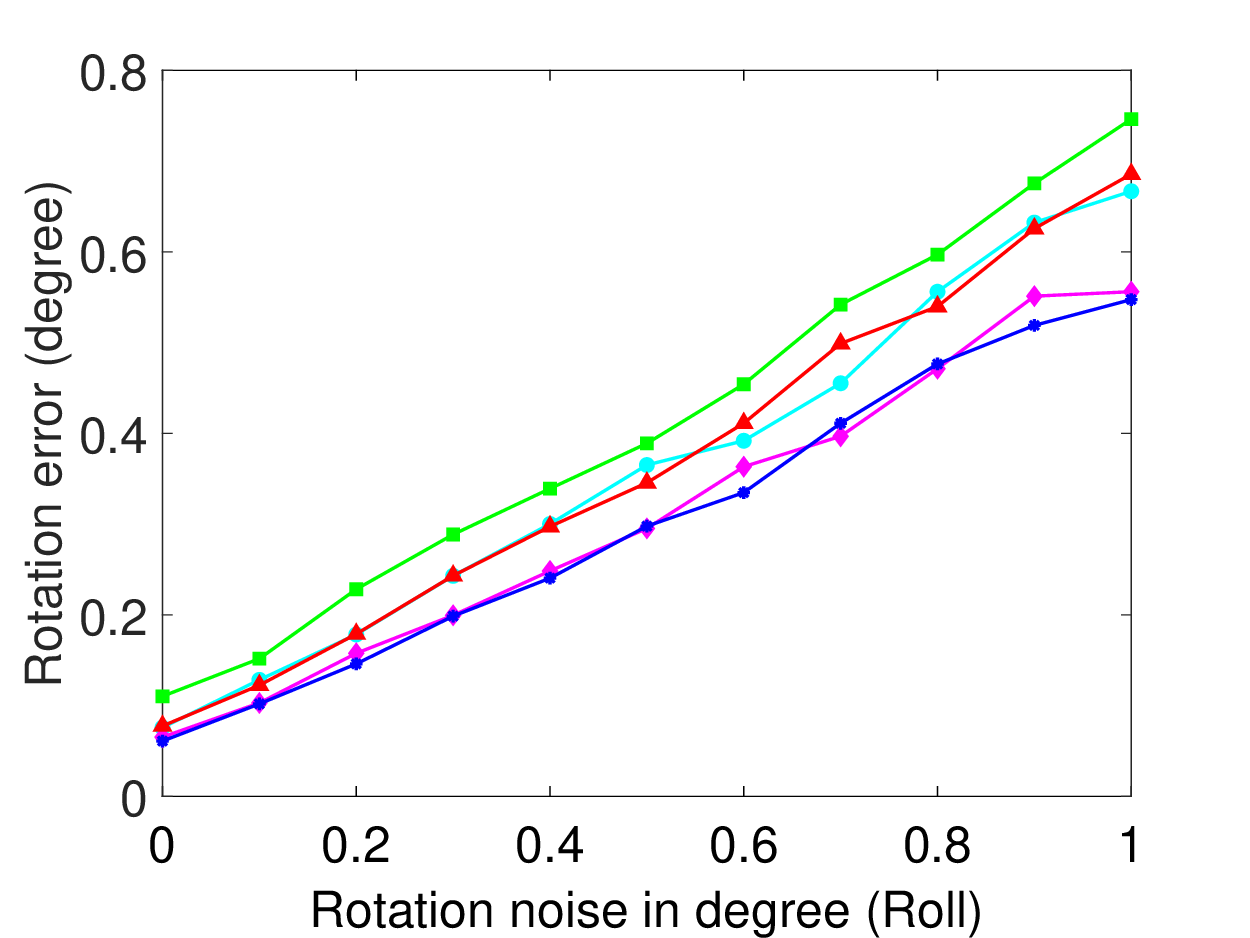}
		}
\subfigure[$\varepsilon_{\mathbf{t}}$]
		{
			\includegraphics[width=0.45\linewidth]{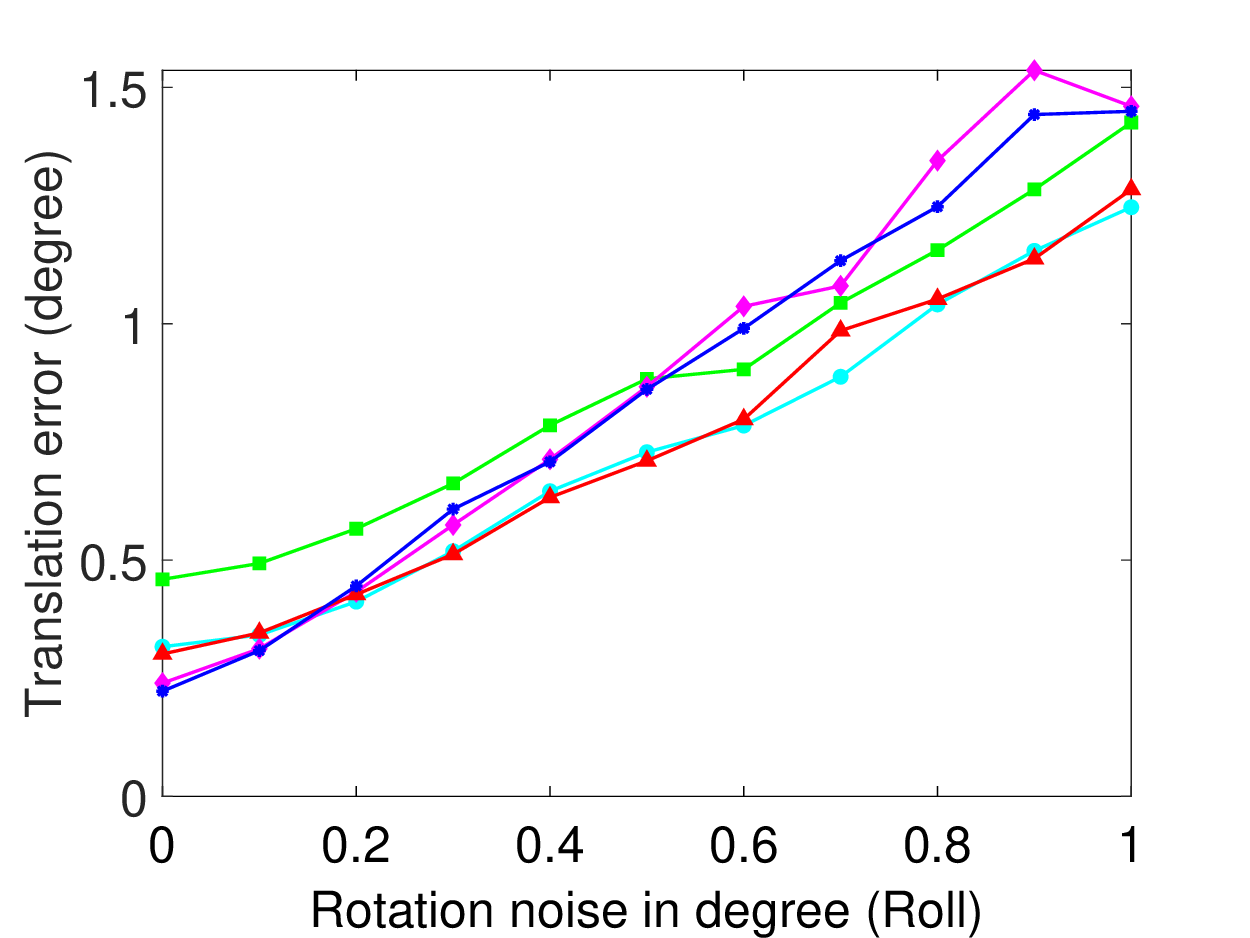}
		}
	 
		\subfigure[${\varepsilon_{\bf{R}}}$]
		{
			\includegraphics[width=0.45\linewidth]{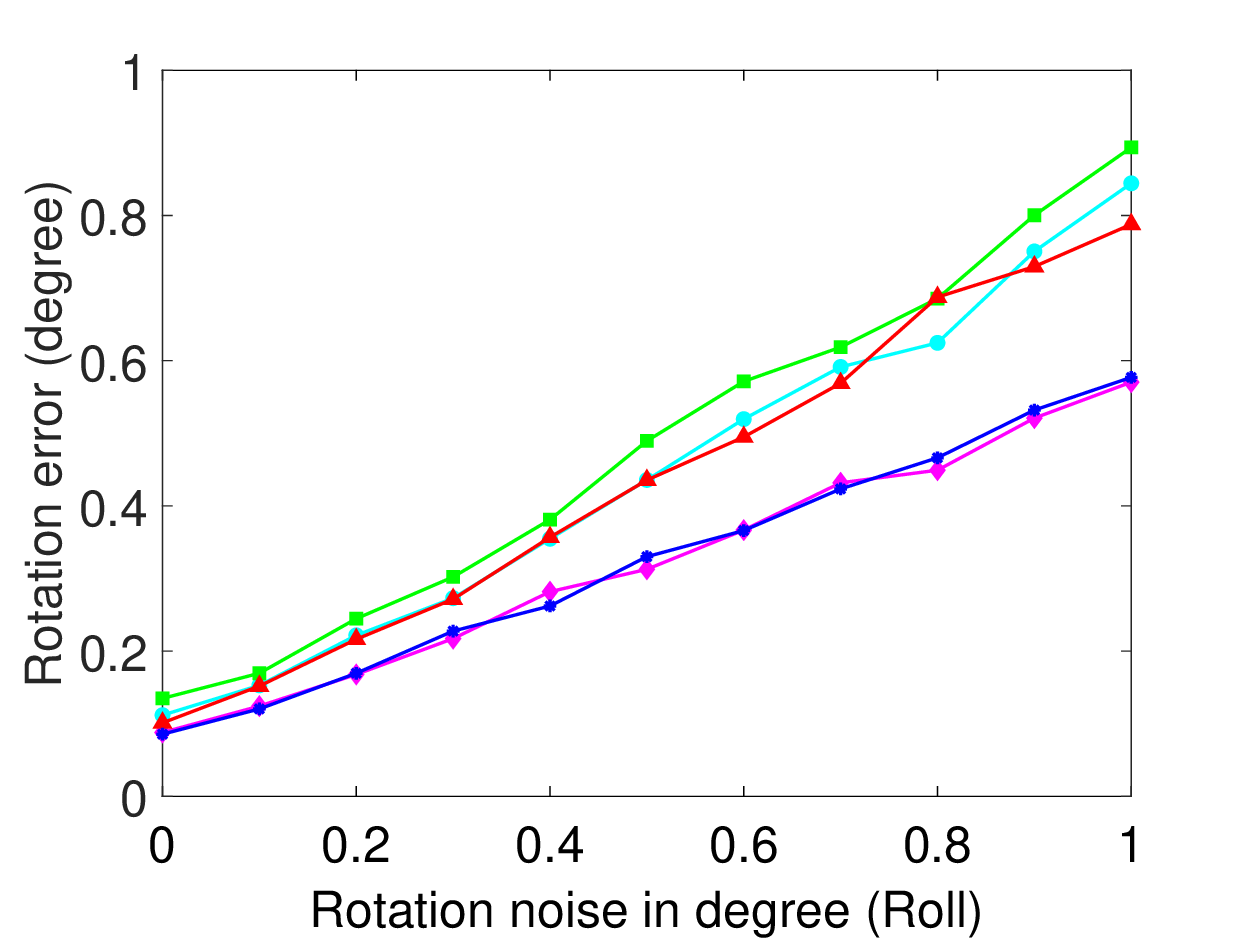}
		}
\subfigure[$\varepsilon_{\mathbf{t}}$]
		{
			\includegraphics[width=0.45\linewidth]{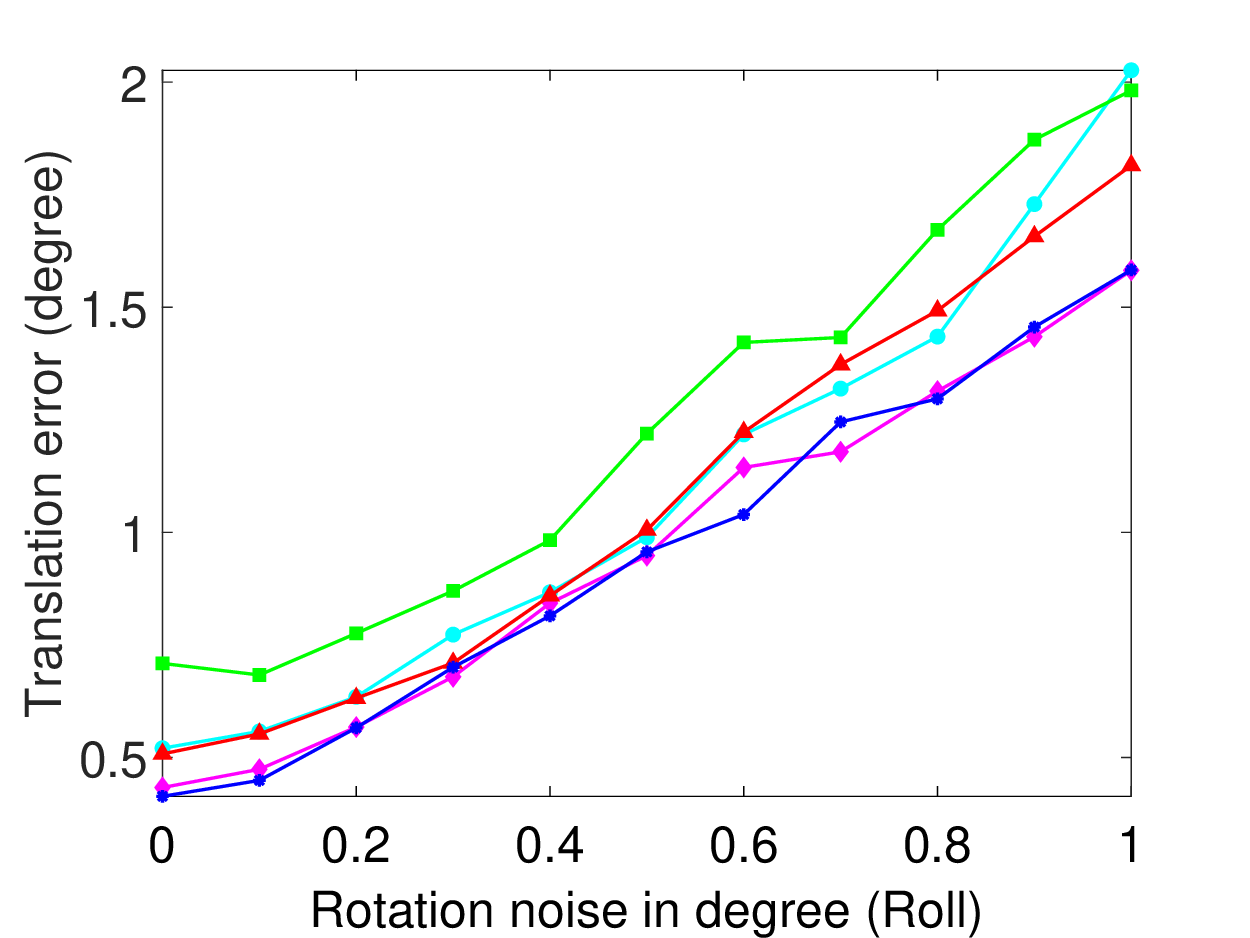}
		}
 
		\subfigure[${\varepsilon_{\bf{R}}}$]
		{
			\includegraphics[width=0.45\linewidth]{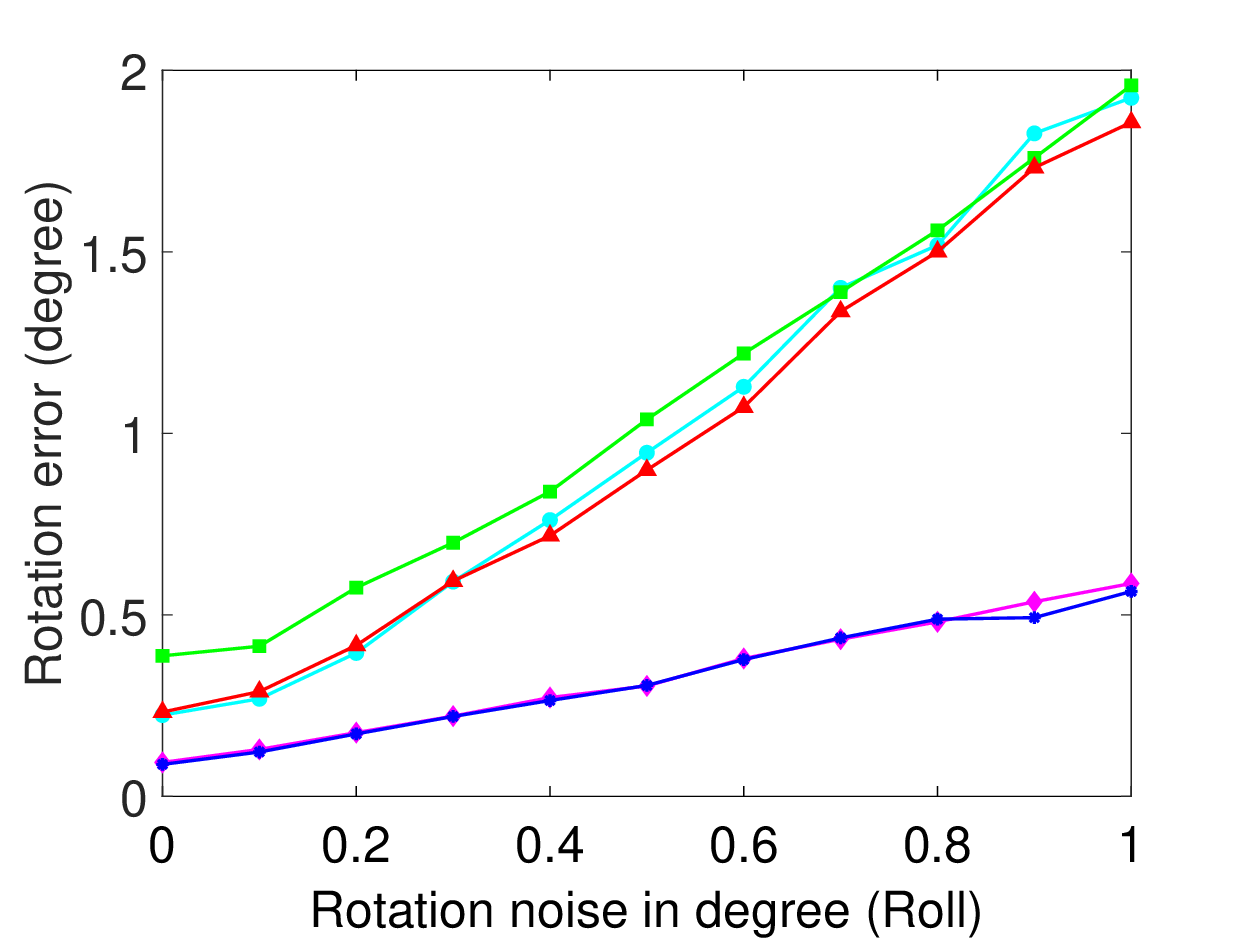}
		}
\subfigure[$\varepsilon_{\mathbf{t}}$]
		{
			\includegraphics[width=0.45\linewidth]{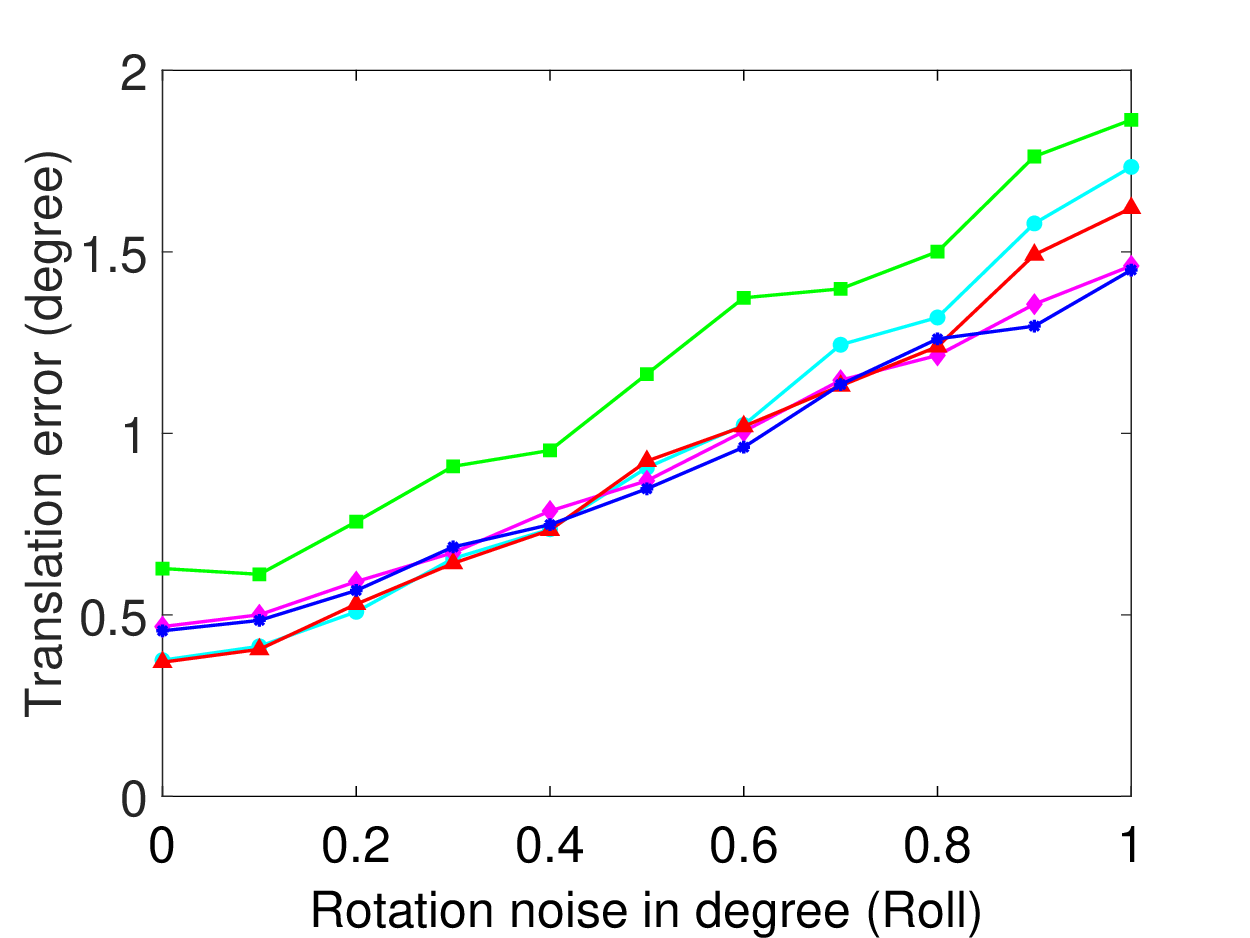}
		}
	\end{center}
	\caption{Rotation and translation errors for multi-camera systems with increasing IMU roll angle noise. The top, middle, and bottom rows present the performance of different solvers under forward, random, and sideways motions, respectively.}
	\label{fig:roll_angle_noise}
\end{figure}

Fig.~\ref{fig:pitch_angle_noise} illustrates the performance of the proposed solvers under increasing pitch angle noise across forward, random, and sideways motion patterns. 
As the pitch angle noise increases, the accuracy of all 4-point solvers gradually degrades.
Among the evaluated methods, \texttt{4pt-Our-Axis} and \texttt{4pt-Sweeney}~\cite{sweeney2014solving} exhibit comparable accuracy and consistently outperform other 4-point solvers. This enhanced performance can be attributed to their rotation representation. Since IMU rotation axis direction noise is generally non-orthogonal to IMU vertical angle noise, this representation appears to effectively mitigate the impact of disturbances caused by IMU vertical angle noise. Meanwhile, \texttt{4pt-Our} and \texttt{4pt-Lee}~\cite{hee2014relative} exhibit similar accuracy and perform better than \texttt{4pt-Liu}~\cite{liu2017robust}. 

Fig.~\ref{fig:roll_angle_noise} illustrates the performance of the proposed solvers under increasing roll angle noise across forward, random, and sideways motion patterns. As the roll angle noise increases, \texttt{4pt-Our-Axis} and \texttt{4pt-Sweeney}~\cite{sweeney2014solving} exhibit comparable performance and achieve better rotation accuracy than the other three methods.
Regarding translation estimation, \texttt{4pt-Our} and \texttt{4pt-Lee}~\cite{hee2014relative} demonstrate similar performance, and both outperform \texttt{4pt-Liu}~\cite{liu2017robust} in forward motion and sideways motion. In random motion, the translation errors of \texttt{4pt-Our} and \texttt{4pt-Our-Axis} are also smaller than those of \texttt{4pt-Liu}~\cite{liu2017robust}.
These results reinforce the robustness of the evaluated methods under diverse angular noise conditions.

\subsection{Experiments on Real Data}
The proposed methods were evaluated on the \texttt{KITTI} dataset~\cite{geiger2013vision}, a widely-used benchmark for evaluating autonomous driving vision algorithms. 
This dataset contains multiple sequences, with ground truth values for sequences 00-10 provided by its onboard GPS/IMU units. We compared the performance across all 11 sequences, which collectively contain approximately 23,000 image pairs.
To ensure robustness against outliers, all methods were embedded within an RANSAC framework. 
 Notably, the final results correspond to the initial hypotheses that maximized the inlier count; no subsequent bundle adjustment or local optimization was performed. 
 Detailed results are presented in Table~\ref{tab:realR} and Table~\ref{tab:realT}. 

 \begin{table}[htbp]
    \caption{ Median rotation errors for KITTI sequences(unit: degree)}
    \label{tab:realR}
        \setlength\tabcolsep{4pt}
    \begin{center}
\begin{tabular}{|c|c|c|c|c|c|c|c|}
\hline

Seq. & \begin{tabular}[c]{@{}c@{}}5pt-Mart.\\ ~\cite{martyushev2020efficient}\end{tabular} & \begin{tabular}[c]{@{}c@{}}4pt-Lee\\ ~\cite{hee2014relative}\end{tabular} & \begin{tabular}[c]{@{}c@{}}4pt-Liu\\ ~\cite{liu2017robust}\end{tabular} & \begin{tabular}[c]{@{}c@{}}4pt-Sw.\\ ~\cite{sweeney2014solving}\end{tabular} & \begin{tabular}[c]{@{}c@{}}4pt-Our\\ \end{tabular} & \begin{tabular}[c]{@{}c@{}}4pt-Our\\ -Axis\end{tabular} \\ \hline 
00 & 0.059 & 0.031 & 0.032 & 0.033 & \textbf{0.031} & 0.032          \\ \hline
01 & 0.047 & 0.044          & 0.043 & 0.028 & 0.045          & \textbf{0.027} \\ \hline
02 & 0.056 & 0.028          & 0.028 & 0.030 & \textbf{0.027} & 0.029          \\ \hline
03 & 0.058 & 0.037          & 0.038 & 0.037 & 0.037          & \textbf{0.036} \\ \hline
04 & 0.036 & \textbf{0.019} & 0.022 & 0.021 & 0.020          & 0.022          \\ \hline
05 & 0.046 & 0.022          & 0.022 & 0.023 & 0.022          & \textbf{0.021} \\ \hline
06 & 0.038 & 0.024          & 0.023 & 0.022 & 0.023          & \textbf{0.021} \\ \hline
07 & 0.047 & 0.022          & 0.022 & 0.022 & \textbf{0.021} & 0.022          \\ \hline
08 & 0.048 & 0.023          & 0.022 & 0.024 & \textbf{0.021} & 0.024          \\ \hline
09 & 0.054 & 0.025          & 0.025 & 0.027 & \textbf{0.024} & 0.026          \\ \hline
10 & 0.057 & 0.024          & 0.024 & 0.026 & \textbf{0.023} & 0.025          \\ \hline
\end{tabular}
    \end{center}
 \end{table}

\begin{table}[htbp]
    \caption{ Median translation direction errors for KITTI sequences(unit: degree)}
    \label{tab:realT}
        \setlength\tabcolsep{4pt}
    \begin{center}
\begin{tabular}{|c|c|c|c|c|c|c|c|}
\hline
Seq. & \begin{tabular}[c]{@{}c@{}}5pt-Mart.\\ ~\cite{martyushev2020efficient}\end{tabular} & \begin{tabular}[c]{@{}c@{}}4pt-Lee\\ ~\cite{hee2014relative}\end{tabular} & \begin{tabular}[c]{@{}c@{}}4pt-Liu\\ ~\cite{liu2017robust}\end{tabular} & \begin{tabular}[c]{@{}c@{}}4pt-Sw.\\ ~\cite{sweeney2014solving}\end{tabular} & \begin{tabular}[c]{@{}c@{}}4pt-Our\\ \end{tabular} & \begin{tabular}[c]{@{}c@{}}4pt-Our\\ -Axis\end{tabular} \\ \hline 

00 & 1.409 & 1.317 & 1.31  & 1.116          & 1.288 & \textbf{1.113} \\ \hline
01 & 1.991 & 2.228 & 2.066 & 1.263          & 2.174 & \textbf{1.218} \\ \hline
02 & 1.112 & 1.062 & 1.068 & 0.977          & 1.069 & \textbf{0.975} \\ \hline
03 & 1.601 & 1.436 & 1.484 & 1.21           & 1.389 & \textbf{1.175} \\ \hline
04 & 0.723 & 0.676 & 0.759 & \textbf{0.605} & 0.756 & 0.612          \\ \hline
05 & 1.113 & 0.928 & 0.925 & 0.776          & 0.923 & \textbf{0.75}  \\ \hline
06 & 0.722 & 0.71  & 0.728 & 0.571          & 0.72  & \textbf{0.566} \\ \hline
07 & 1.463 & 1.066 & 1.064 & \textbf{0.918} & 1.039 & 0.95           \\ \hline
08 & 1.477 & 1.412 & 1.371 & 1.248          & 1.372 & \textbf{1.222} \\ \hline
09 & 0.995 & 0.922 & 0.894 & 0.873          & 0.933 & \textbf{0.824} \\ \hline
10 & 1.101 & 1.05  & 1.059 & 0.92           & 1.065 & \textbf{0.883} \\ \hline
\end{tabular}
    \end{center}
 \end{table}

\begin{figure}[htbp]              
     \subfigure[4pt-Lee]{
     \includegraphics[width=0.45\linewidth]{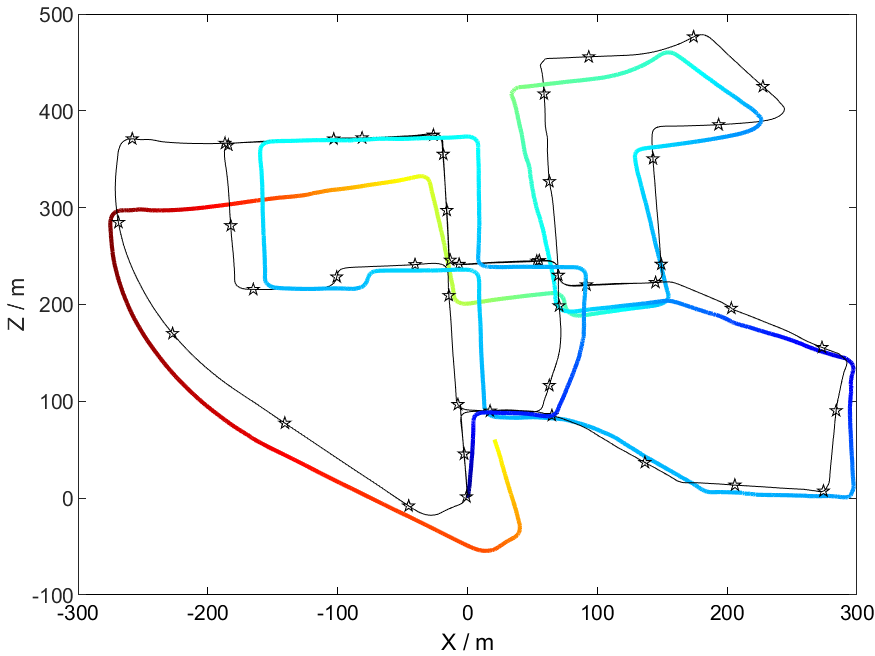}
     \label{fig.guiji a}
      }%
   \subfigure[4pt-Liu]{
     \includegraphics[width=0.45\linewidth]{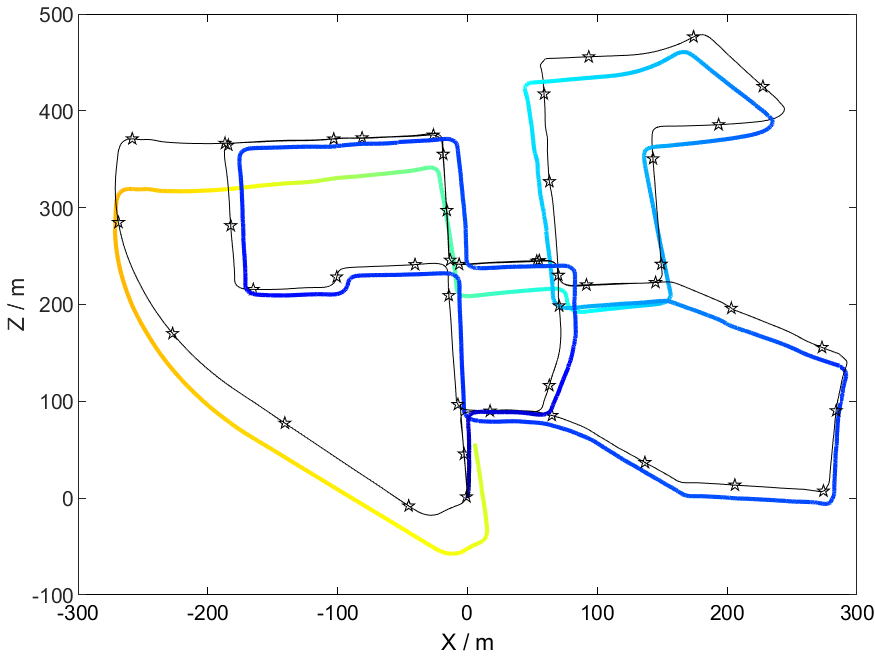}
     \label{fig.guiji b}
      }%
\\
     \subfigure[4pt-Sweeney]{
     \includegraphics[width=0.45\linewidth]{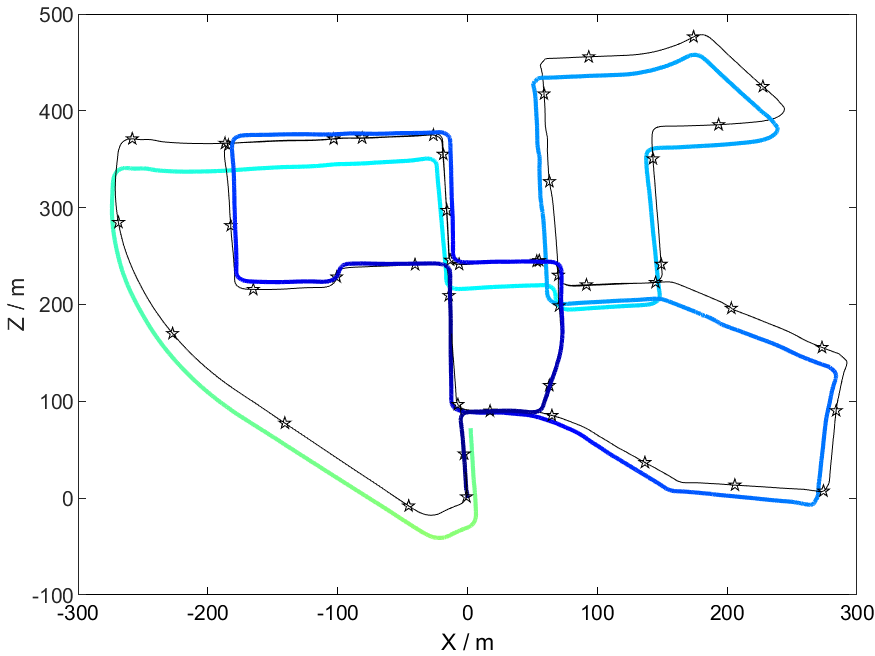}
     \label{fig.guiji c}
      }%
   \subfigure[4pt-Our]{
     \includegraphics[width=0.45\linewidth]{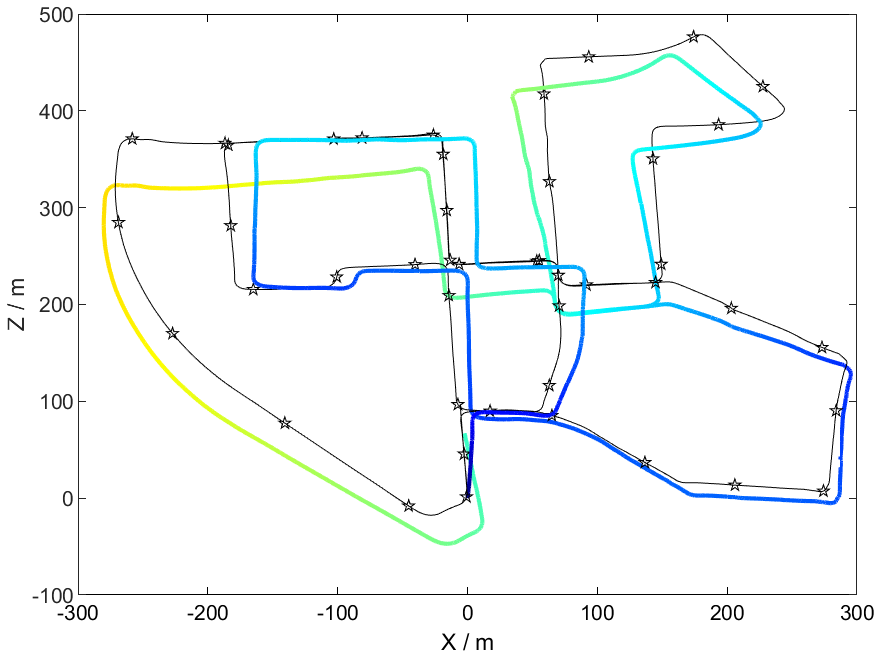}
     \label{fig.guiji d}
      }%
\\
           \centering
    \subfigure[4pt-Our-Axis]{
     \centering
     \includegraphics[width=0.50\linewidth]{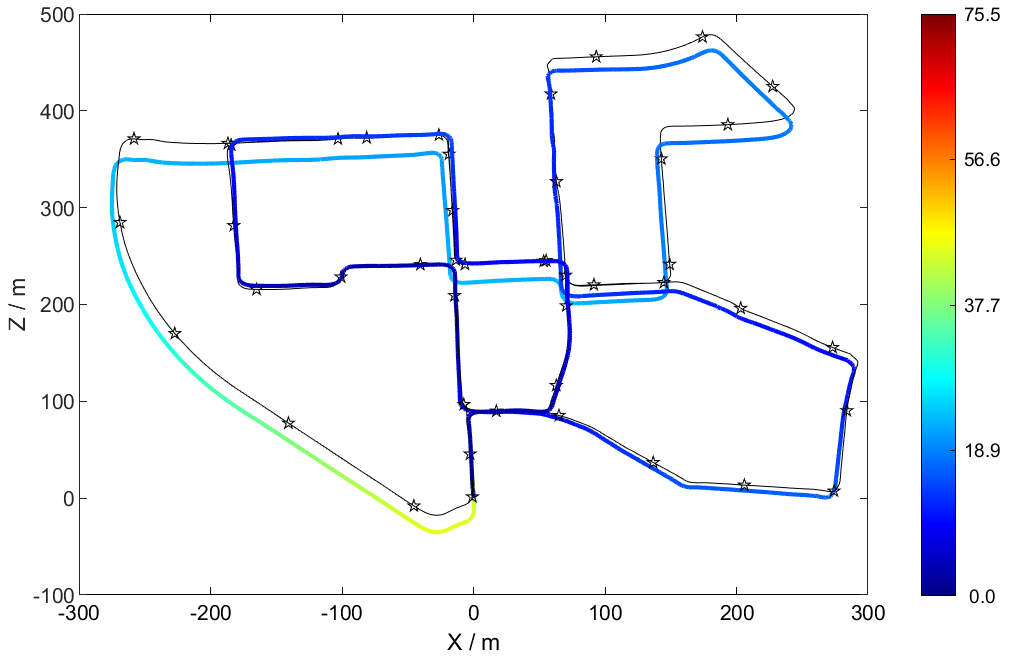}
     \label{fig.guiji e}
      }%
    
    \caption{ Estimated visual odometry trajectories for KITTI sequences 00 (unit: meter). (a) Trajectory estimation with \texttt{4pt-Lee} method, (b) Trajectory estimation with \texttt{4pt-Liu} method, (c) Trajectory estimation with \texttt{4pt-Sweeney} method, (d) Trajectory estimation with \texttt{4pt-Our} method, (e) Trajectory estimation with \texttt{4pt-Our-Axis} method.}

    \label{fig:GUIJI}
     \vspace{-10pt}
\end{figure}

From Table~\ref{tab:realR} and Table~\ref{tab:realT}, it can be observed that the 4-point solvers consistently achieve higher rotation accuracy than the 5-point solver \texttt{5pt-Martyushev}~\cite{martyushev2020efficient} across all 11 sequences.  
In terms of translation accuracy, the 4-point methods also outperform the 5-point solver in most sequences, with the exception of sequences 01, 04, and 06. 
Within the 4-point solvers, our proposed methods demonstrate a slight advantage. Notably, \texttt{4pt-Our-Axis} achieves the best translation accuracy in all sequences except sequences 04 and 07.
Since our proposed methods fall within the category of 4-point solvers, we provide a more intuitive comparison among these methods by visualizing the estimated camera trajectories in Fig.~\ref{fig:GUIJI}. The ground truth trajectory is shown in black, while the estimated trajectories are color-coded according to their absolute trajectory error (ATE) magnitudes \cite{sturm2012benchmark}. For conciseness, we present only the trajectory results for sequence 00. As shown in Fig.~\ref{fig:GUIJI}, the \texttt{4pt-Our-Axis} and \texttt{4pt-Sweeney} solver achieves the highest accuracy compared to other methods. Furthermore, among the 4-point solvers utilizing IMU vertical angle priors, \texttt{4pt-Our} also exhibits smaller ATE values than \texttt{4pt-Lee} and \texttt{4pt-Liu}. These observations further validate the effectiveness of our proposed method in real-world settings.

\section{conclusion}
In this work, we present two efficient minimal solvers for estimating the relative pose of multi-camera systems using a minimal number of four point correspondences. The first solver leverages the vertical direction provided by IMUs, while the second utilizes the rotation axis direction prior. Both proposed solvers are built upon a novel depth-based parameterization of translation, as opposed to traditional translation representations. Building on this foundation, we reduce the 4-DOF multi-camera relative pose estimation problem to solving a univariate 6th-degree polynomial. Extensive simulations and real-world experiments demonstrate that our solvers achieve improved computational efficiency while maintaining competitive accuracy compared to state-of-the-art solvers for multi-camera ego-motion estimation.

\bibliographystyle{IEEEtran}
\bibliography{references}
 \vspace{-15pt}

\end{document}